%% file: paper.tex
\documentclass{article}
\pdfoutput=1
\usepackage[nolist]{acronym} 
\usepackage[usenames, dvipsnames]{color} 
\usepackage{enumerate, footmisc, verbatim, subfigure, amsmath, amsthm, amssymb, fullpage, csquotes, dashrule, tikz, bbm, booktabs, bm}
\usepackage[framemethod=TikZ]{mdframed}
\usepackage[numbers]{natbib}

\usepackage[pdftex,
            pdfauthor={David Abel, D. Ellis Hershkowitz, Michael L. Littman},
            pdftitle={Near Optimal Behavior via Approximate State Abstraction},
            pdfkeywords={Reinforcement Learning, Abstraction, State Aggregation, State Abstraction, Planning, Markov Decision Process, MDP}]{hyperref}

\input{commands}

\begin{document}

\title{\Large{ {\bf Near Optimal Behavior via Approximate State Abstraction}}\blfootnote{A previous version of this paper was published in the Proceedings of the 33rd International Conference on Machine Learning, New York, NY, USA, 2016. JMLR: W\&CP volume 48. Copyright 2016 by the author(s).}}

\date{}

\author{
  \normalsize{ {\bf David Abel}\footnote{The first two authors contributed equally.}} \\ \normalsize{Brown University} \\ \normalsize{\texttt{david\_abel@brown.edu}}
  \and
  \normalsize{ {\bf D. Ellis Hershkowitz}$^\dagger$} \\ \normalsize{Carnegie Mellon University} \\ \normalsize{\texttt{dhershko@cs.cmu.edu}}
  \and
  \normalsize{ {\bf Michael L. Littman}} \\ \normalsize{Brown University} \\ \normalsize{\texttt{mlittman@cs.brown.edu}}
}

\maketitle

\begin{acronym}
\acro{MDP}{Markov Decision Process}
\acrodefplural{MDP}[MDPs]{Markov Decision Processes}
\acro{RL}{reinforcement learning}
\acrodefplural{RL}{Reinforcement learning}
\end{acronym}

\begin{abstract}
\input{abstract.tex}
\end{abstract}

\input{introduction.tex}

\input{mdp_background.tex}

\input{related_work.tex}

\input{abstraction_notation.tex}

\input{theory_results.tex}
\input{example_domains.tex}

\input{results_discussion.tex}

\input{conclusion.tex}

\bibliographystyle{plainnat}
\bibliography{abstraction}

\end{document}

%% file: commands.tex
\definecolor{dblue}{RGB}{98, 140, 190}
\definecolor{dlblue}{RGB}{216, 235, 255}
\definecolor{dgreen}{RGB}{124, 155, 127}
\definecolor{dpink}{RGB}{207, 166, 208}
\definecolor{dyellow}{RGB}{255, 248, 199}
\definecolor{dgray}{RGB}{46, 49, 49}

\newcommand*\circled[1]{\tikz[baseline=(char.base)]{\node[shape=circle,draw,inner sep=0.7pt] (char) {\footnotesize{#1}};}}

\newcommand{\eps}{\varepsilon}

\newcommand{\mc}[1]{\mathcal{#1}}

\newmdenv[
  topline=false,
  bottomline=false,
  rightline = false,
  leftmargin=10pt,
  rightmargin=0pt,
  innertopmargin=0pt,
  innerbottommargin=0pt
]{innerproof}

\newcounter{DaveThmCounter}
\newcounter{DaveDefCounter}
\newtheorem{thm}{Theorem}
\newtheorem{lma}{Lemma}
\newtheorem{clm}{Claim}

\newtheoremstyle{dotless}{}{}{\itshape}{}{\bfseries}{}{ }{}
\theoremstyle{dotless}
\newtheorem{defn}{Definition}

\newcommand{\bdefn}[1]{
\begin{defn} {\boldmath (#1)}\textbf{:}
}

\newcommand{\edefn}[1]{
\end{defn}
}

\newcommand{\ep}{\widetilde \phi}
\newcommand{\epQ}{\ep_{Q^*,\varepsilon}}
\newcommand{\epB}{\ep_{\text{bolt},\varepsilon}}
\newcommand{\epMu}{\ep_{\text{mult},\varepsilon}}
\newcommand{\epM}{\ep_{\text{model},\varepsilon}}

\DeclareMathOperator*{\argmax}{arg\,max}
\newcommand{\mcS}{\mathcal{S}}
\newcommand{\mcT}{\mathcal{T}}
\newcommand{\mcR}{\mathcal{R}}
\newcommand{\mcA}{\mathcal{A}}

\renewcommand{\thefootnote}{\fnsymbol{footnote}}

\makeatletter
\def\blfootnote{\gdef\@thefnmark{}\@footnotetext}
\addtocounter{footnote}{1}%
\makeatother

%% file: abstract.tex
%
The combinatorial explosion that plagues planning and \ac{RL} algorithms can be moderated using state abstraction. Prohibitively large task representations can be condensed such that essential information is preserved, and consequently, solutions are tractably computable. However, exact abstractions, which treat only fully-identical situations as equivalent, fail to present opportunities for abstraction in environments where no two situations are exactly alike. In this work, we investigate approximate state abstractions, which treat nearly-identical situations as equivalent. We present theoretical guarantees of the quality of behaviors derived from four types of approximate abstractions. Additionally, we empirically demonstrate that approximate abstractions lead to reduction in task complexity and bounded loss of optimality of behavior in a variety of environments. 


%% file: introduction.tex
\section{Introduction}
\label{sec:intro}

Abstraction plays a fundamental role in learning. Through abstraction, intelligent agents may reason about only the salient features of their environment while ignoring what is irrelevant. Consequently, agents are able to solve considerably more complex problems than they would be able to without the use of abstraction. However, \textit{exact abstractions}, which treat only fully-identical situations as equivalent, require complete knowledge that is computationally intractable to obtain. Furthermore, often no two situations are identical, so exact abstractions are often ineffective. To overcome these issues, we investigate \textit{approximate abstractions} that enable agents to treat sufficiently similar situations as identical. This work characterizes the impact of equating ``sufficiently similar'' states in the context of planning and \ac{RL} in \acp{MDP}. The remainder of our introduction contextualizes these intuitions in \acp{MDP}.

\begin{figure}[h]
\centering
\includegraphics[width=0.52\columnwidth]{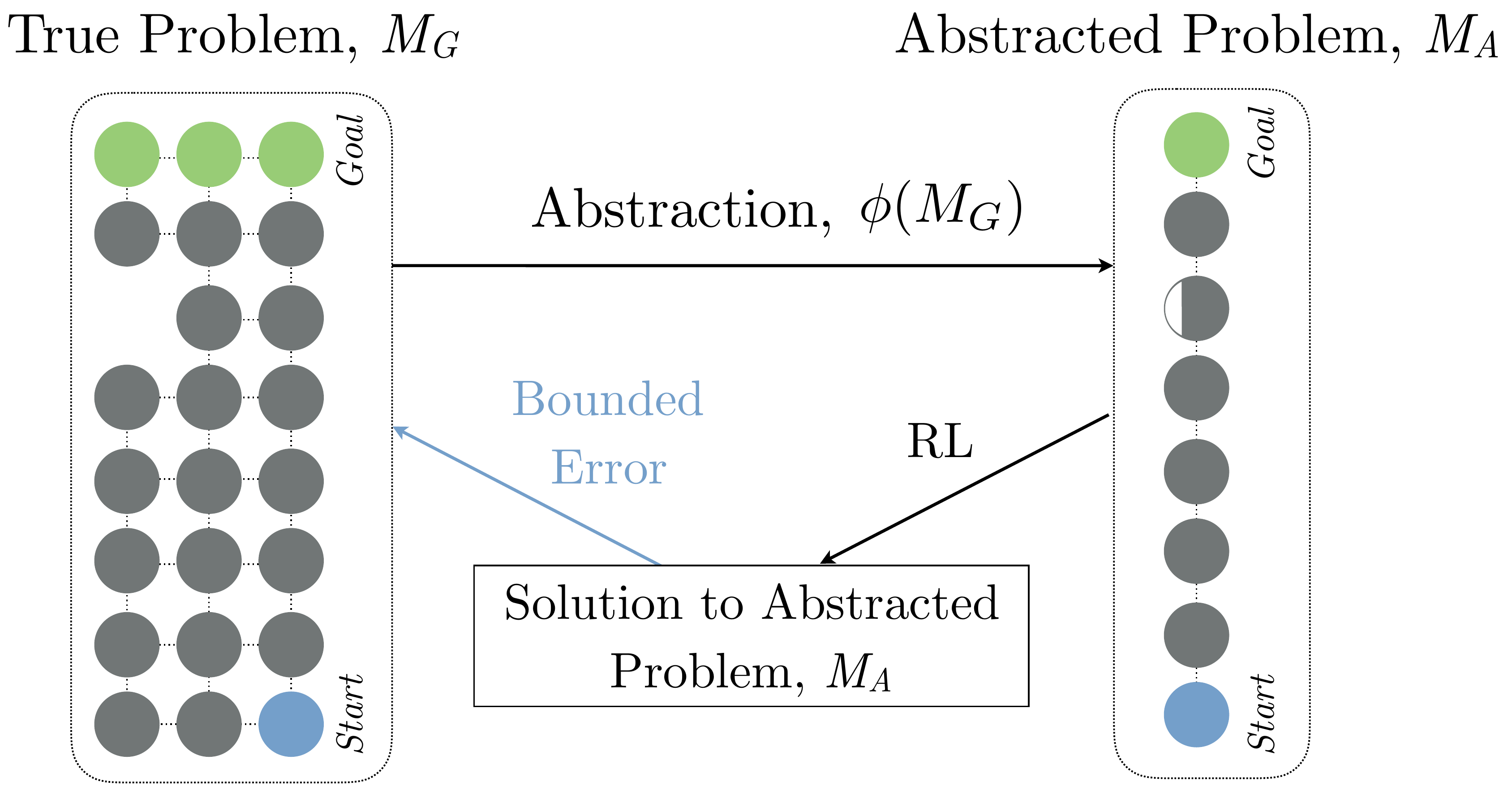}
\label{fig:main}
\caption{We investigate families of approximate state abstraction functions that induce abstract MDP's whose optimal policies have bounded value in the original MDP.}
\end{figure}

Solving for optimal behavior in \acp{MDP} in a planning setting is known to be P-Complete in the size of the state space~\cite{papadimitriou1987complexity,littman1995complexity}. Similarly, many \ac{RL} algorithms for solving \acp{MDP} are known to require a number of samples polynomial in the size of the state space~\cite{Strehl2009}. Although polynomial runtime or sample complexity may seem like a reasonable constraint, the size of the state space of an \ac{MDP} grows super-polynomially with the number of variables that characterize the domain - a result of Bellman's curse of dimensionality. Thus, solutions polynomial in state space size are often ineffective for sufficiently complex tasks. For instance, a robot involved in a pick-and-place task might be able to employ planning algorithms to solve for how to manipulate some objects into a desired configuration in time polynomial in the number of states, but the number of states it must consider grows exponentially with the number of objects with which it is working \cite{abel2015goal}.

Thus, a key research agenda for planning and \ac{RL} is leveraging abstraction to reduce large state spaces~\cite{andre2002state,jong2005state,dearden1997abstraction,dietterich2000hierarchical,Bean2011}. This agenda has given rise to methods that reduce \textit{ground} \acp{MDP} with large state spaces to \textit{abstract} MDPs with smaller state spaces by aggregating states according to some notion of equality or similarity. In the context of \acp{MDP}, we understand exact abstractions as those that aggregate states with equal values of particular quantities, for example, optimal $Q$-values. Existing work has characterized how exact abstractions can fully maintain optimality in \acp{MDP}~\cite{li2006towards,dean1997modelmin}. 

\renewcommand*{\thefootnote}{\arabic{footnote}}

The thesis of this work is that performing approximate abstraction in \acp{MDP} by relaxing the state aggregation criteria from equality to similarity achieves polynomially bounded error in the resulting behavior while offering three benefits. First, approximate abstractions employ the sort of knowledge that we expect a planning or learning algorithm to compute without fully solving the \ac{MDP}. In contrast, exact abstractions often require solving for optimal behavior, thereby defeating the purpose of abstraction. Second, because of their relaxed criteria, approximate abstractions can achieve greater degrees of compression than exact abstractions. This difference is particularly important in environments where no two states are identical. Third, because the state aggregation criteria are relaxed to near equality, approximate abstractions are able to tune the aggressiveness of abstraction by adjusting what they consider sufficiently similar states. 

We support this thesis by describing four different types of approximate abstraction functions that preserve near-optimal behavior by aggregating states on different criteria: $\epQ$, on similar optimal $Q$-values, $\epM$, on similarity of rewards and transitions, $\epB$, on similarity of a Boltzmann distribution over optimal $Q$-values, and $\epMu$, on similarity of a multinomial distribution over optimal $Q$-values. Furthermore, we empirically demonstrate the relationship between the degree of compression and error incurred on a variety of \acp{MDP}.

This paper is organized as follows. In the next section, we introduce the necessary terminology and background of \acp{MDP} and state abstraction. Section~\ref{sec:rel_work} surveys existing work on state abstraction applied to sequential decision making. Section~\ref{sec:theory_results} introduces our primary result; bounds on the error guaranteed by four classes of approximate state abstraction. The following two sections introduce simulated domains used in experiments (Section~\ref{sec:example_domains}), and a discussion of experiments in which we apply one class of approximate abstraction to a variety of different tasks to empirically illustrate the relationship between degree of compression and error incurred (Section~\ref{sec:results}).

%% file: mdp_background.tex
\section{\acp{MDP} and Sequential Decision Making}
\label{sec:background}



An \ac{MDP} is a problem representation for sequential decision making agents, represented by a five-tuple: $\langle \mathcal{S}, \mathcal{A}, \mathcal{T}, \mathcal{R}, \gamma \rangle$. Here, $\mathcal{S}$ is a finite state space; $\mathcal{A}$ is a finite set of actions available to the agent; $\mathcal{T}$ denotes $\mathcal{T}(s,a, s')$, the probability of an agent transitioning to state $s' \in \mathcal{S}$ after applying action $a \in \mathcal{A}$ in state $s \in \mathcal{S}$; $\mathcal{R}(s,a)$ denotes the reward received by the agent for executing action $a$ in state $s$; $\gamma \in [0, 1]$ is a discount factor that determines how much the agent prefers future rewards over immediate rewards. We assume without loss of generality that the range of all reward functions is normalized to $[0,1]$. The solution to an \ac{MDP} is called a policy, denoted $\pi: \mathcal{S} \mapsto \mathcal{A}$.

The objective of an agent is to solve for the policy that maximizes its expected discounted reward from any state, denoted $\pi^*$. We denote the expected discounted reward for following policy $\pi$ from state $s$ as the value of the state under that policy, $V^\pi(s)$. We similarly denote the expected discounted reward for taking action $a \in \mathcal{A}$ and then following policy $\pi$ from state $s$ forever after as $Q^\pi(s,a)$, defined by the Bellman Equation as:
\begin{equation}
Q^\pi(s,a) = \mathcal{R}(s,a) + \gamma \sum_{s'} \mathcal{T}(s,a,s') Q^\pi(s',\pi(s')).
\end{equation}
We let \textsc{RMax} denote the maximum reward (which is 1), and \textsc{QMax} denote the maximum $Q$ value, which is $\frac{\textsc{RMax}}{1-\gamma}$. The value function, $V$, defined under a given policy, denoted $V^\pi(s)$, is defined as:
\begin{equation}
V^\pi(s) = Q^\pi(s,\pi(s)).
\end{equation}
Lastly, we denote the value and $Q$ functions under the optimal policy as $V^*$ or $V^{\pi^*}$ and $Q^*$ or $Q^{\pi^*}$, respectively. For further background, see~\citet{kaelbling1996reinforcement}.

%% file: related_work.tex
\section{Related Work}
\label{sec:rel_work}

Several other projects have addressed similar topics.

\subsection{Approximate State Abstraction}
\citet{dean1997model} leverage the notion of {\it bisimulation} to investigate partitioning an \ac{MDP}'s state space into clusters of states whose transition model and reward function are within $\varepsilon$ of each other. They develop an algorithm called Interval Value Iteration (IVI) that converges to the correct bounds on a family of abstract MDPs called Bounded \acp{MDP}.

Several approaches build on \citet{dean1997model}. \citet{ferns2004metrics,ferns2006methods} investigated state similarity metrics for \acp{MDP}; they bounded the value difference of ground states and abstract states for several bisimulation metrics that induce an abstract MDP. This differs from our work which develops a theory of abstraction that bounds the suboptimality of applying the optimal policy of an abstract MDP to its ground MDP, covering four types of state abstraction, one of which closely parallels bisimulation. \citet{even2003approximate} analyzed different distance metrics used in identifying state space partitions subject to $\varepsilon$-similarity, also providing value bounds (their Lemma 4) for $\eps$-homogeneity subject to the $L_\infty$ norm, which parallels our Claim 2. \citet{ortner2013adaptive} developed an algorithm for learning partitions in an online setting by taking advantage of the confidence bounds for $\mcT$ and $\mcR$ provided by UCRL~\cite{auer2009near}.

\citet{hutter2016extreme,hutter2014extreme} investigates state aggregation beyond the MDP setting. Hutter presents a variety of results for aggregation functions in reinforcement learning. Most relevant to our investigation is Hutter's Theorem 8, which illustrates properties of aggregating states based on similar $Q$ values. Hutter's Theorem part (a) parallels our Claim: both bound the value difference between ground and abstraction states, and part (b) is analogous to our Lemma 1: both bound the value difference of applying the optimal abstraction policy in the ground, and part (c) is a repetition of the comment given by~\citet{li2006towards} that $Q^*$ abstractions preserve the optimal value function. For Lemma 1, our proof strategies differ from Hutter's, but the result is the same.

Approximate state abstraction has also been applied to the planning problem, in which the agent is given a model of its environment and must compute a plan that satisfies some goal. \citet{Hostetler2014} apply state abstraction to Monte Carlo Tree Search and expectimax search, giving value bounds of applying the optimal abstract action in the ground tree(s), similarly to our setting.~\citet{dearden1997abstraction} also formalize state-abstraction for planning, focusing on abstractions that are quickly computed and offer bounded value. Their primary analysis is on abstractions that remove negligible literals from the planning domain description, yielding value bounds for these abstractions and a means of incrementally improving abstract solutions to planning problems.~\citet{jiang2014improving} analyze a similar setting, applying abstractions to the Upper Confidence Bound applied to Trees algorithm adapted for planning, introduced by~\citet{kocsis2006bandit}.

~\citet{mandelefficient} advance Bayesian aggregation in RL to define Thompson Clustering for Reinforcement Learning (TCRL), an extension of which achieves near-optimal Bayesian regret bounds. ~\citet{Jiang2015} analyze the problem of choosing between two candidate abstractions. They develop an algorithm based on statistical tests that trades of the approximation error with the estimation error of the two abstractions, yielding a loss bound on the quality of the chosen policy.

\subsection{Specific Abstraction Algorithms}
Many previous works have targeted the creation of algorithms that enable state abstraction for MDPs. \citet{andre2002state} investigated a method for state abstraction in hierarchical reinforcement learning leveraging a programming language called ALISP that promotes the notion of {\it safe} state abstraction. Agents programmed using ALISP can ignore irrelevant parts of the state, achieving abstractions that maintain optimality. \citet{dietterich2000hierarchical} developed MAXQ, a framework for composing tasks into an abstracted hierarchy where state aggregation can be applied.~\citet{bakker2004hierarchical} also target hierarchical abstraction, focusing on subgoal discovery. \citet{jong2005state} introduced a method called {\it policy-irrelevance} in which agents identify (online) which state variables may be safely abstracted away in a factored-state \ac{MDP}.~\citet{Dayan1993} develop ``Feudal Reinforcement Learning" which presents an early form of hierarchical RL that restructures $Q$-Learning to manage the decomposition of a task into subtasks. For a more complete survey of algorithms that leverage state abstraction in past reinforcement-learning papers, see \citet{li2006towards}, and for a survey of early works on hierarchical reinforcement learning, see \citet{barto2003recent}.

\subsection{Exact Abstraction Framework}

\citet{li2006towards} developed a framework for exact state abstraction in \acp{MDP}. In particular, the authors defined five types of state aggregation functions, inspired by existing methods for state aggregation in \acp{MDP}. We generalize two of these five types, $\phi_{Q^*}$ and $\phi_{\text{model}}$, to the approximate abstraction case. Our generalizations are equivalent to theirs when exact criteria are used (i.e. $\varepsilon = 0$). Additionally, when exact criteria are used our bounds indicate that no value is lost, which is one of core results of \citet{li2006towards}.~\citet{walsh2006transferring} build on the framework they previously developed by showing empirically how to transfer abstractions between structurally related MDPs.

%% file: abstraction_notation.tex
\section{Abstraction Notation}
\label{sec:abs_not}

We build upon the notation used by \citet{li2006towards}, who introduced a unifying theoretical framework for state abstraction in \acp{MDP}.

\bdefn{$M_G$, $M_A$}
We understand an abstraction as a mapping from the state space of a ground \ac{MDP}, $M_G$, to that of an abstract MDP, $M_A$, using a state aggregation scheme. Consequently, this mapping induces an abstract \ac{MDP}. Let $M_G = \langle \mathcal{S}_G, \mathcal{A}, \mathcal{T}_G, \mathcal{R}_G, \gamma \rangle$ and $M_A = \langle \mathcal{S}_A, \mathcal{A}, \mathcal{T}_A, \mathcal{R}_A, \gamma \rangle$.
\edefn

\bdefn{$\mcS_A$, $\phi$}
The states in the abstract \ac{MDP} are constructed by applying a state aggregation function, $\phi$, to the states in the ground \ac{MDP}, $\mathcal{S}_A$. More specifically, $\phi$ maps a state in the ground \ac{MDP} to a state in the abstract \ac{MDP}:
\begin{equation}
\mathcal{S}_A = \{ \phi(s) \mid s \in \mathcal{S}_G\}.
\end{equation}
\edefn


\bdefn{$G$}
Given a $\phi$, each ground state has associated with it the ground states with which it is aggregated. Similarly, each abstract state has its constituent ground states. We let $G$ be the function that retrieves these states:
\begin{equation}
G(s)=
\begin{cases}
\{g \in \mcS_G \mid \phi(g) = \phi(s) \},& \text{if } s \in \mcS_G,\\
\{g \in \mcS_G \mid \phi(g)=s \},& \text{if } s \in \mcS_A.
\end{cases}
\end{equation}
\edefn

The abstract reward function and abstract transition dynamics for each abstract state are a weighted combination of the rewards and transitions for each ground state in the abstract state.


\bdefn{$\omega(s)$}
We refer to the weight associated with a ground state, $s \in \mathcal{S}_G$ by $\omega(s)$. The only restriction placed on the weighting scheme is that it induces a probability distribution on the ground states of each abstract state:
\begin{equation}
\forall s \in \mcS_G \left(\sum_{s \in G(s)} \omega(s)\right) = 1 \hspace{3mm} \textrm{ AND }\hspace{3mm}  \omega(s) \in [0,1].
\end{equation}
\edefn

\bdefn{$\mcR_A$}
The abstract reward function $\mathcal{R}_A: \mathcal{S}_A \times \mathcal{A} \mapsto [0,1]$ is a weighted sum of the rewards of each of the ground states that map to the same abstract state:
\begin{equation}
\mathcal{R}_A(s,a) = \sum_{g \in G(s)} \mathcal{R}_G(g,a) \omega(g) .
\end{equation}
\edefn

\bdefn{$\mcT_A$}
The abstract transition function $\mathcal{T}_A: \mathcal{S}_A \times \mathcal{A} \times \mathcal{S}_A \mapsto [0,1]$ is a weighted sum of the transitions of each of the ground states that map to the same abstract state:
\begin{equation}
\mathcal{T}_A(s,a,s') = \sum_{g \in G(s)} \sum_{g' \in G(s')} \mathcal{T}_G(g,a,g') \omega(g).
\end{equation}
\edefn

%% file: theory_results.tex
\section{Approximate State Abstraction}
\label{sec:theory_results}

Here, we introduce our formal analysis of approximate state abstraction, including results bounding the error associated with these abstraction methods. In particular, we demonstrate that abstractions based on approximate $Q^*$ similarity (\ref{sec:Q*}), approximate model similarity (\ref{sec:model}), and approximate similarity between distributions over $Q^*$, for both Boltzmann (\ref{sec:boltz}) and multinomial (\ref{sec:mult}) distributions induce abstract \acp{MDP} for which the optimal policy has bounded error in the ground MDP.

We first introduce some additional notation.

\bdefn{$\pi_A^*$, $\pi_G^*$}
We let $\pi_A^* : \mcS_A \rightarrow \mcA$ and $\pi_G^* : \mcS_G \rightarrow \mcA$ stand for the optimal policies in the abstract and ground \acp{MDP}, respectively.
\edefn

We are interested in how the optimal policy in the abstract \ac{MDP} performs in the ground \ac{MDP}. As such, we formally define the policy in the ground \ac{MDP} derived from optimal behavior in the abstract \ac{MDP}:

\bdefn{$\pi_{GA}$}
Given a state $s \in \mcS_G$ and a state aggregation function, $\phi$, 
\begin{equation}
\pi_{GA}(s)=\pi_A^*(\phi(s)).
\end{equation}
\edefn

We now define types of abstraction based on functions of state--action pairs.
\bdefn{$\ep_{f,\varepsilon}$}
Given a function $f: \mathcal{S}_G \times \mathcal{A} \rightarrow \mathbb{R}$ and a fixed non-negative $\varepsilon \in \mathbb{R}$, we define $\ep_{f,\varepsilon}$ as a type of approximate state aggregation function that satisfies the following for any two ground states $s_1$, $s_2$: 
\begin{equation}
\label{eq:phi_f}
\ep_{f,\varepsilon}(s_1) = \ep_{f,\varepsilon}(s_2) \rightarrow \forall_a \left|f(s_1, a) - f(s_2, a)\right| \leq \varepsilon.
\end{equation}
\edefn

\noindent That is, when $\ep_{f,\varepsilon}$ aggregates states, all aggregated states have values of $f$ within $\varepsilon$ of each other for all actions. \\

Finally, we estliabsh notation to distinguish between the {\it ground} and {\it abstract} value ($V$) and action value ($Q$) functions.

\bdefn{$Q_G$, $V_G$}
Let $Q_G = Q^{\pi_G^*} : \mathcal{S}_G \times \mathcal{A} \rightarrow \mathbb{R} $ and $V_G = V^{\pi_G^*}: \mathcal{S}_G \rightarrow \mathbb{R} $ denote the optimal Q and optimal value functions in the ground \ac{MDP}.
\edefn

\bdefn{$Q_A$, $V_A$}
Let $Q_A  = Q^{\pi_A^*}: \mathcal{S}_A \times \mathcal{A} \rightarrow \mathbb{R}$ and $V_A  = V^{\pi_A^*}: \mathcal{S}_A \rightarrow \mathbb{R}$  stand for the optimal Q and optimal value functions in the abstract \ac{MDP}.
\edefn

\subsection{Main Result}
We now introduce the main result of the paper.

\newpage
\begin{mdframed}[roundcorner=1pt, backgroundcolor=white]
\begin{thm}
There exist at least four types of approximate state aggregation functions, $\epQ$, $\epM$, $\epB$ and $\epMu$, for which the optimal policy in the abstract \ac{MDP}, applied to the ground \ac{MDP}, has suboptimality bounded polynomially in ${\varepsilon}$:
\begin{equation}
\forall_{s \in \mathcal{S}_G} V_G^{\pi_G^*}(s) - V_G^{\pi_{GA}}(s) \leq 2 \eps \eta_f
\label{eq:main_result}
\end{equation}

\noindent Where $\eta_f$ differs between abstraction function families:
\begin{align*}
\eta_{Q^*} &= \frac{1}{(1-\gamma)^2} \\
\eta_{\text{model}} &= \frac{1 + \gamma \left( |\mc{S}_G| - 1 \right)}{(1-\gamma)^3} \\
\eta_{\text{bolt}} &= \frac{\left(\frac{|\mc{A}|}{1-\gamma} + \eps k_{\text{bolt}} + k_{\text{bolt}}\right)}{(1-\gamma)^2} \\
\eta_{\text{mult}} &= \frac{\left(\frac{|\mc{A}|}{1-\gamma} + k_{\text{mult}}\right)}{(1-\gamma)^2}
%
\end{align*}
\end{thm}
\end{mdframed}

\vspace{4mm}

\noindent For $\eta_{\text{bolt}}$ and $\eta_{mult}$, we also assume that the difference in the normalizing terms of each distribution is bounded by some non-negative constant, $k_{\text{mult}}, k_{\text{bolt}} \in \mathbb{R}$, of $\varepsilon$:
\begin{align*}
\left |\sum_i Q_G(s_1,a_i) - \sum_j Q_G(s_2,a_j) \right | &\leq k_{\text{mult}} \times \varepsilon \\
\left |\sum_i e^{Q_G(s_1,a_i)} - \sum_j e^{Q_G(s_2,a_j)} \right | &\leq k_{\text{bolt}} \times \varepsilon
\end{align*}
Naturally, the value bound of Equation~\ref{eq:main_result} is meaningless for $2 \eps \eta_f \geq \frac{\textsc{RMax}}{1-\gamma} = \frac{1}{1-\gamma}$, since this is the maximum possible value in any MDP (and we assumed the range of $\mc{R}$ is $[0,1]$). In light of this, observe that for $\eps = 0$, all of the above bounds are exactly 0. Any value of $\eps$ interpolated between these two points achieves different degrees of abstraction, with different degrees of bounded loss.

We now introduce each approximate aggregation family and prove the theorem by proving the specific value bound for each function type.

\subsection{Optimal Q Function: $\ep_{Q^*,\varepsilon}$}
\label{sec:Q*}

We consider an approximate version of \citet{li2006towards}'s $\phi_{Q^*}$. In our abstraction, states are aggregated together when their optimal $Q$-values are within $\varepsilon$.
\bdefn{$\epQ$}
An approximate $Q$ function abstraction has the same form as Equation~\ref{eq:phi_f}:
\vspace{-1mm}
\begin{equation}
\ep_{Q^*,\varepsilon}(s_1) = \ep_{Q^*,\varepsilon}(s_2) \rightarrow \forall_a \left|Q_G(s_1, a) - Q_G(s_2, a)\right| \leq \varepsilon.
\end{equation}
\edefn

\begin{lma}
\label{lma:Q*}
When a $\ep_{Q^*,\varepsilon}$ type abstraction is used to create the abstract \ac{MDP}:
\vspace{-2mm}
\begin{equation}
\forall_{s \in \mcS_G} V_G^{\pi_G^*}(s) - V_G^{\pi_{GA}}(s) \leq \frac{2\varepsilon}{(1-\gamma)^2}.
\end{equation}
\vspace{-4mm}
\end{lma}
\noindent \textbf{Proof of Lemma~\ref{lma:Q*}:}
We first demonstrate that $Q$-values in the abstract \ac{MDP} are close to $Q$-values in the ground \ac{MDP} (Claim \ref{clm:closeQs}). We next leverage Claim \ref{clm:closeQs} to demonstrate that the optimal action in the abstract \ac{MDP} is nearly optimal in the ground \ac{MDP}  (Claim \ref{clm:optAbsActionNearOptGround}). Lastly, we use Claim \ref{clm:optAbsActionNearOptGround} to conclude Lemma \ref{lma:Q*} (Claim \ref{clm:lmaFromClm}). \\

\begin{clm}
\label{clm:closeQs}
Optimal $Q$-values in the abstract \ac{MDP} closely resemble optimal $Q$-values in the ground \ac{MDP}:
\begin{equation}
\label{eq:Q*Claim1}
\forall_{s_G \in \mathcal{S}_G, a} |Q_G(s_G, a) - Q_A(\epQ(s_G), a)| \leq \frac{\varepsilon}{1-\gamma}.
\end{equation}
\end{clm}

Consider a non-Markovian decision process of the same form as an \ac{MDP}, $M_T = \langle \mathcal{S}_T, \mathcal{A}_G, \mathcal{R}_T, \mathcal{T}_T, \gamma \rangle$, parameterized by integer an $T$, such that for the first $T$ time steps the reward function, transition dynamics and state space are those of the abstract MDP, $M_A$, and after $T$ time steps the reward function, transition dynamics and state spaces are those of $M_G$. Thus,
\begin{align*}
\mathcal{S}_T &= \begin{cases}
\mathcal{S}_G& \text{if } T = 0 \\
\mathcal{S}_A& \text{o/w}
\end{cases} \\
\mathcal{R}_T(s,a) &= \begin{cases}
\mathcal{R}_G(s,a)& \text{if } T = 0\\
\mathcal{R}_A(s, a)& \text{o/w}
\end{cases}\\
\mcT_T(s,a,s') &= \begin{cases}
\mcT_G(s,a,s')& \text{if }T = 0\\
\underset{{g \in G(s)}}{\sum}\left[\mcT_G(g, a, s') \omega(g) \right]& \text{if } T = 1\\
\mcT_A(s,a,s')& \text{o/w}
\end{cases}
\end{align*}
The $Q$-value of state $s$ in $\mathcal{S}_T$ for action $a$ is:
\begin{equation}
Q_T(s, a) = 
\begin{cases}
	   Q_G(s, a) &  \text{if } T=0\\
	   \underset{g \in G(s)}{\sum} \left[ Q_G(g,a) \omega(g) \right] & \text{if } T = 1\\
	   \mathcal{R}_A(s,a) + \sigma_{T-1}(s,a) &\text{o/w}
\end{cases}
\end{equation}
where:
\begin{equation*}
\sigma_{T-1}(s,a) = \gamma \underset{{s_A}' \in \mathcal{S}_A}{\sum} \mathcal{T}_A(s,a,{s_A}') \max_{a'} Q_{T-1}({s_A}', a').
\end{equation*}
We proceed by induction on $T$ to show that:
\begin{equation}
\label{eq:clm1Induct}
\forall_{T, s_G \in \mathcal{S}_G, a} |Q_T(s_T, a) - Q_G(s_G, a) | \leq \sum_{t=0}^{T-1} \varepsilon \gamma^{t},
\end{equation}
where $s_T = s_G$ if $T=0$ and $s_T = \epQ(s_G)$ otherwise. \\

\noindent \textit{Base Case: $T = 0$} \\

\noindent When $T = 0$, $Q_T = Q_G$, so this base case trivially follows. \\

\noindent \textit{Base Case: $T = 1$} \\

\noindent By definition of $Q_T$, we have that $Q_1$ is
\begin{align*}
&Q_1(s,a) = \underset{g \in G(s)}{\sum} \left[ Q_G(g,a) \omega(g) \right].
\end{align*}
Since all co-aggregated states have $Q$-values within $\varepsilon$ of one another and $\omega(g)$ induces a convex combination,
\begin{align*}
&Q_1(s_T,a) \leq \varepsilon \gamma^t + \varepsilon + Q_G(s_G, a) \\
\therefore& \left| Q_{1}(s_T, a) - Q_G(s_G,a) \right| \leq \sum_{t=0}^{1}\varepsilon \gamma^t.
\end{align*}
\noindent \textit{Inductive Case: $T > 1$} \\

\noindent We assume as our inductive hypothesis that:
\begin{equation*}
\forall_{s_G \in \mathcal{S}_G, a} |Q_{T-1}(s_T, a) - Q_G(s_G, a) | \leq \sum_{t=0}^{T-2} \varepsilon \gamma^t.
\end{equation*}

\noindent Consider a fixed but arbitrary state, $s_G \in \mathcal{S}_G$, and fixed but arbitrary action $a$.
Since $T > 1$, $s_T$ is $\epQ(s_G)$.
By definition of $Q_{T}(s_T, a)$, $\mathcal{R}_A$, $\mathcal{T}_A$:
\[
Q_T(s_T, a) = \sum_{g \in G(s_T)}\omega(g)\ \times \left[ \mathcal{R}_G(g,a) + \gamma \sum_{g' \in \mathcal{S}_G} \mathcal{T}_G(g,a,g') \max_{a'} Q_{T-1}(g', a')      \right].
\]
Applying our inductive hypothesis yields:
\[
Q_T(s_T, a) \leq \sum_{g \in G(s_T)}\omega(g) \times \biggl[ R_G(g,a)\ + \gamma \sum_{g' \in \mathcal{S}_G} T_G(g,a,g') \max_{a'}(Q_G(g', a') + \sum_{t=0}^{T-2} \varepsilon \gamma^t) \biggr].
\]
Since all aggregated states have $Q$-values within $\varepsilon$ of one another:
\begin{align*}
Q_T(s_T, a) \leq \gamma\sum_{t=0}^{T-2} \varepsilon \gamma^t + \varepsilon + Q_G(s_G, a).
\end{align*}
Since $s_G$ is arbitrary we conclude Equation \ref{eq:clm1Induct}. As $T \rightarrow \infty$, $\sum_{t=0}^{T-1} \varepsilon \gamma^t \rightarrow \frac{\varepsilon}{1-\gamma}$ by the sum of infinite geometric series and $Q_T \rightarrow Q_A$. Thus, Equation \ref{eq:clm1Induct} yields Claim \ref{clm:closeQs}.

\begin{clm}
\label{clm:optAbsActionNearOptGround}

Consider a fixed but arbitrary state, $s_G \in \mathcal{S}_G$ and its corresponding abstract state $s_A=\epQ(s_G)$.
Let $a^*_G$ stand for the optimal action in $s_G$, and $a^*_A$ stand for the optimal action in $s_A$:
\begin{align*}
a^*_G = \argmax_a Q_G(s_G, a), \hspace{4mm}
a^*_A = \argmax_a Q_A(s_A, a).
\end{align*}
The optimal action in the abstract MDP has a $Q$-value in the ground MDP that is nearly optimal:
\begin{equation}
\label{eq:Q*Claim2}
V_G(s_G) \leq Q_G(s_G, a^*_A) + \frac{2\varepsilon}{1-\gamma}.
\end{equation}
\end{clm}
\noindent By Claim~\ref{clm:closeQs},
\begin{align}
&V_G(s_G) = Q_G(s_G, a^*_G) \leq Q_A(s_A, a^*_G) + \frac{\varepsilon}{1-\gamma}.
\label{eq:Q*OptActionResult}
\end{align}
By the definition of $a^*_A$, we know that 
\begin{align}
Q_A(s_A, a^*_G) + \frac{\varepsilon}{1-\gamma} \leq Q_A(s_A, a^*_A) + \frac{\varepsilon}{1-\gamma}.
\end{align}
Lastly, again by Claim~\ref{clm:closeQs}, we know
\begin{align}
Q_A(s_A, a^*_A) + \frac{\varepsilon}{1-\gamma} \leq Q_G(s_g, a^*_A) + \frac{2\varepsilon}{1-\gamma}.
\end{align}
Therefore, Equation~\ref{eq:Q*Claim2} follows.

\begin{clm}
Lemma \ref{lma:Q*} follows from Claim \ref{clm:optAbsActionNearOptGround}.
\label{clm:lmaFromClm}
\end{clm}

\noindent Consider the policy for $M_G$ of following the optimal abstract policy $\pi^*_A$ for $t$ steps and then following the optimal ground policy $\pi^*_G$ in $M_G$:
\begin{equation}
\pi_{A,t}(s)=
\begin{cases}
\pi_G^*(s) &\text{if } t= 0\\
\pi_{GA}(s) &\text{if } t > 0
\end{cases}
\end{equation}

\noindent For $t > 0$, the value of this policy for $s_G \in \mathcal{S}_G$ in the ground \ac{MDP} is:
\[
V_G^{\pi_{A,t}}(s_G) = R_G(s, \pi_{A,t}(s_G)) +\ \gamma \sum_{{s_G}' \in \mathcal{S}_G}\mathcal{T}_G(s_G, a, {s_G}')V_G^{\pi_{A,t-1}}({s_G}').
\]

\noindent For $t=0$, $V_G^{\pi_{A,t}}(s_G)$ is simply $V_G(s_G)$.

\noindent We now show by induction on $t$ that
\begin{equation}
\forall_{t, s_G \in \mathcal{S}_g} V_G(s_G) \leq  V_G^{\pi_{A,t}}(s_G) + \sum_{i=0}^{t}\gamma^i \frac{2\varepsilon}{1-\gamma}.
\end{equation}

\noindent \textit{Base Case: $t=0$} \\

\noindent By definition, when $t=0$, $V_G^{\pi_{A,t}} = V_G$, so our bound trivially holds in this case. \\

\noindent \textit{Inductive Case: $t > 0$} \\

\noindent Consider a fixed but arbitrary state $s_G \in \mathcal{S}_G$.
We assume for our inductive hypothesis that
\begin{equation}
V_G(s_G) \leq V_G^{\pi_{A,t-1}}(s_G) + \sum_{i=0}^{t-1}\gamma^i \frac{2\varepsilon}{1-\gamma}.
\end{equation}
By definition,
\[
V_G^{\pi_{A,t}}(s_G) = R_G(s, \pi_{A,t}(s_G)) + \gamma \sum_{g'}\mathcal{T}_G(s_G, a, {s_G}')V_G^{\pi_{A,t-1}}({s_G}').
\]
Applying our inductive hypothesis yields:
\[
V_G^{\pi_{A,t}}(s_G) \geq R_G(s_G, \pi_{A,t}(s_G)) + \gamma \sum_{{s_G}'}\mathcal{T}_G(s_G, \pi_{A,t}(s_G), {s_G}')\left(V_G({s_G}') - \sum_{i=0}^{t-1}\gamma^i \frac{2\varepsilon}{1-\gamma} \right).
\]
Therefore,
\begin{align*}
V_G^{\pi_{A,t}}(s_G) &\geq -\gamma\sum_{i=0}^{t-1}\gamma^i \frac{2\varepsilon}{1-\gamma} + Q_G(s_G, \pi_{A,t} (s_G)).
\end{align*}
Applying Claim~\ref{clm:optAbsActionNearOptGround} yields:
\begin{align*}
&V_G^{\pi_{A,t}}(s_G) \geq -\gamma\sum_{i=0}^{t-1}\gamma^i \frac{2\varepsilon}{1-\gamma} - \frac{2\varepsilon}{1-\gamma} + V_{G}(s_G) \\
\therefore\ &V_G(s_G) \leq V_G^{\pi_{A,t}}(s_G)  + \sum_{i=0}^{t}\gamma^i \frac{2\varepsilon}{1-\gamma}.
\end{align*}
Since $s_G$ was arbitrary, we conclude that our bound holds for all states in $\mathcal{S}_G$ for the inductive case.
Thus, from our base case and induction, we conclude that
\begin{equation}
\forall_{t, s_G \in \mathcal{S}_g} V_G^{\pi_G^*}(s_G) \leq  V_G^{\pi_{A,t}}(s_G) + \sum_{i=0}^{t}\gamma^i \frac{2\varepsilon}{1-\gamma}.
\end{equation}

\noindent Note that as $t \rightarrow \infty$, $\sum_{i=0}^{t}\gamma^i \frac{2\varepsilon}{1-\gamma} \rightarrow \frac{2\varepsilon}{(1-\gamma)^2}$ by the sum of infinite geometric series and $\pi_{A,t}(s) \rightarrow \pi_{GA}$.
Thus, we conclude Lemma~\ref{lma:Q*}.
\qed

\subsection{Model Similarity: $\ep_{model,\varepsilon}$}
\label{sec:model}

Now, consider an approximate version of \citet{li2006towards}'s $\phi_{model}$, where states are aggregated together when their rewards and transitions are within $\varepsilon$.
\bdefn{$\epM$}
We let $\epM$ define a type of abstraction that, for fixed $\varepsilon$, satisfies:
\begin{multline}
\epM(s_1) = \epM(s_2) \rightarrow \\
\forall_a \left| \mcR_G(s_1, a) - \mcR_G(s_2, a)\right| \leq \varepsilon\; \text{~AND}\ \
\forall_{s_A \in \mcS_A} \left|\sum_{{s_G}' \in G(s_A)} \left[\mcT_G(s_1, a, {s_G}') - \mcT_G(s_2, a,{s_G}')\right] \right| \leq \varepsilon.
\end{multline}
\edefn

\begin{lma}
\label{lma:model}
When $\mcS_A$ is created using a $\ep_{model,\varepsilon}$ type:
\begin{equation}
\forall_{s \in \mcS_G} V_G^{\pi_G^*}(s) - V_G^{\pi_{GA}}(s) \leq \frac{2\eps + 2\gamma \eps \left( |\mc{S}_G| - 1 \right)}{(1-\gamma)^3}.
\end{equation}
\end{lma}

\noindent {\bf Proof of Lemma~\ref{lma:model}:} \\

\noindent Let $B$ be the maximum $Q$-value difference between any pair of ground states in the same abstract state for $\epM$:
\begin{equation*}
B = \max_{s_1, s_2, a}  |Q_G(s_1, a) - Q_G(s_2, a)|,
\end{equation*}
where $s_1, s_2 \in G(s_A)$. First, we expand:
\begin{equation}
B=\max_{s_1, s_2, a}      \biggl|\mcR_G(s_1, a) - \mcR_G(s_2, a)\ + \gamma \sum_{{s_G}' \in \mcS_G} \biggl[(\mcT_G(s_1,a,{s_G}') -\mcT_G(s_2, a, {s_G}'))\max_{a'}Q_G({s_G}', a')\biggr]\biggr|
\end{equation}
Since difference of rewards is bounded by $\varepsilon$:
\begin{equation}
B\leq \varepsilon + \gamma \sum_{s_A \in \mcS_A}\sum_{{s_G}' \in G(s_A)} \biggl[(T_G(s_1, a, {s_G}')\ - T_G(s_2, a, {s_G}')) \max_{a'}Q_G({s_G}', a') \biggr].
\end{equation}
By similarity of transitions under $\epM$:
\begin{align*}
B \leq \varepsilon + \gamma \textsc{QMax} \sum_{s_A \in \mcS_A} \varepsilon \leq \varepsilon + \gamma|\mcS_G|\varepsilon \textsc{QMax}.
\end{align*}
Recall that \textsc{QMax} $= \frac{\textsc{RMax}}{1-\gamma}$, and we defined $\textsc{RMax} = 1$:
\begin{equation*}
B \leq \frac{\varepsilon + \gamma(|\mcS_G| - 1) \varepsilon}{1-\gamma}.
\end{equation*}
Since the $Q$-values of ground states grouped under $\epM$ are strictly less than $B$, we can understand $\epM$ as a type of $\ep_{Q^*,B}$. Applying Lemma \ref{lma:Q*} yields Lemma \ref{lma:model}.
\qed

\subsection{Boltzmann over Optimal Q: $\epB$}
\label{sec:boltz}

Here, we introduce $\epB$, which aggregates states with similar Boltzmann distributions on $Q$-values. This type of abstractions is appealing as Boltzmman distributions balance exploration and exploitation~\cite{sutton1998reinforcement}. We find this type particularly interesting for abstraction purposes as, unlike $\epQ$, it allows for aggregation when $Q$-value ratios are similar but their magnitudes are different.

\bdefn{$\epB$}
We let $\epB$ define a type of abstractions that, for fixed $\varepsilon$, satisfies:
\begin{equation}
\epB(s_1) = \epB(s_2) \rightarrow \forall_{a} \left|\frac{e^{Q_G(s_1,a)}}{\sum_b e^{Q_G(s_1,b)}} - \frac{e^{Q_G(s_2,a)}}{\sum_b e^{Q_G(s_2,b)}}\right| \leq \varepsilon.
\label{eq:phi_bolt}
\end{equation}
\edefn

\noindent We also assume that the difference in normalizing terms is bounded by some non-negative constant, $k_{\text{bolt}}  \in \mathbb{R}$, of $\varepsilon$:
\begin{equation}
\left| \sum_b e^{Q_G(s_1,b)} - \sum_b e^{Q_G(s_2,b)} \right| \leq k_{\text{bolt}} \times\varepsilon.
\label{eq:bolt_denom}
\end{equation}
\begin{lma} When $S_A$ is created using a function of the $\epB$ type, for some non-negative constant $k \in \mathbb{R}$:
\begin{equation}
\forall_{s \in \mcS_G} V_G^{\pi^*_G}(s) - V_G^{\pi_{GA}}(s) \leq \frac{2\varepsilon\left(\frac{|\mathcal{A}|}{1-\gamma} + \varepsilon k_{\text{bolt}}  + k_{\text{bolt}} \right)}{(1-\gamma)^2}.
\end{equation}
\label{lma:bolt_lemma}
\end{lma}
\vspace{-3mm}
\noindent We use the approximation for $e^x$, with $\delta$ error:
\begin{equation}
 e^x = 1 + x + \delta  \approx 1 + x.
\label{eq:e_to_x_approx}
\end{equation}
We let $\delta_1$ denote the error in approximating $e^{Q_G(s_1,a)}$ and $\delta_2$ denote the error in approximating $e^{Q_G(s_2,a)}$. \\

\noindent {\bf Proof of Lemma~\ref{lma:bolt_lemma}:} \\

\noindent By the approximation in Equation~\ref{eq:e_to_x_approx} and the assumption in Equation~\ref{eq:bolt_denom}:
\begin{align}
\left|\frac{1 + Q_G(s_1,a) + \delta_1}{\sum_j e^{Q_G(s_1,a_j)}} - \frac{1 + Q_G(s_2,a) + \delta_2}{\sum_j e^{Q_G(s_1,a_j)} \underbrace{\pm k\varepsilon}_{\circled{a}}}\right| \leq \varepsilon \label{eq:bolt_with_approx}
\end{align}
Either term \circled{a} is positive or negative. First suppose the former. It follows by algebra that:
\begin{equation}
-\varepsilon \leq \frac{1 + Q_G(s_1,a) + \delta_1}{\sum_j e^{Q_G(s_1,a_j)}} - \frac{1 + Q_G(s_2,a) + \delta_2}{\sum_j e^{Q_G(s_1,a_j)} + \varepsilon k_{\text{bolt}} } \leq \varepsilon
\end{equation}
Moving terms:
\begin{multline}
-\varepsilon \left(k\varepsilon + \sum_j e^{Q_G(s_1,a_j)}\right) - \delta_1 + \delta_2 \leq \\
\varepsilon k_{\text{bolt}} \left(\frac{1+Q_G(s_1,a) + \delta_1}{\sum_j e^{Q_G(s_1,a_j)}}\right) + Q_G(s_1,a) - Q_G(s_2,a) \leq \\
\varepsilon \left(\varepsilon k_{\text{bolt}}  + \sum_j e^{Q_G(s_1,a_j)}\right) - \delta_1 + \delta_2
\label{eq:a_p_case}
\end{multline}
When \circled{a} is the negative case, it follows that:
\begin{equation}
-\varepsilon \leq \frac{1 + Q_G(s_1,a) + \delta_1}{\sum_j e^{Q_G(s_1,a_j)}} - \frac{1 + Q_G(s_2,a) + \delta_2}{\sum_j e^{Q_G(s_1,a_j)} - \varepsilon k_{\text{bolt}} } \leq \varepsilon
\end{equation}

\noindent By similar algebra that yielded Equation~\ref{eq:a_p_case}:
\begin{multline}
-\varepsilon \left(-\varepsilon k_{\text{bolt}}  + \sum_j e^{Q_G(s_1,a_j)}\right) - \delta_1 + \delta_2 \leq \\
-k\varepsilon\left(\frac{1+Q_G(s_1,a) + \delta_1}{\sum_j e^{Q_G(s_1,a_j)}}\right) + Q_G(s_1,a) - Q_G(s_2,a) \leq \\
\varepsilon \left(\varepsilon k_{\text{bolt}}  + \sum_j e^{Q_G(s_1,a_j)}\right) - \delta_1 + \delta_2
\label{eq:a_m_case}
\end{multline}

\noindent Combining Equation~\ref{eq:a_p_case} and Equation~\ref{eq:a_m_case} results in:
\begin{equation}
\left|Q_G(s_1,a) - Q_G(s_2,a)\right| \leq \varepsilon \left(\frac{|\mathcal{A}|}{1-\gamma} + \varepsilon k_{\text{bolt}}  + k_{\text{bolt}}  \right).
\label{eq:bolt_qs}
\end{equation}
Consequently, we can consider $\epB$ as a special case of the $\ep_{Q^*,B}$ type, where $B = \varepsilon \left(\frac{|\mathcal{A}|}{1-\gamma} + \varepsilon k_{\text{bolt}}  + k_{\text{bolt}}  \right)$. Lemma~\ref{lma:bolt_lemma} then follows from Lemma~\ref{lma:Q*}.
\qed

\subsection{Multinomial over Optimal Q: $\epMu$}
\label{sec:mult}

We consider approximate abstractions derived from a multinomial distribution over $Q^*$ for similar reasons to the Boltzmann distribution. Additionally, the multinomial distribution is appealing for its simplicity.
\bdefn{$\epMu$}
We let $\epMu$ define a type of abstraction that, for fixed $\varepsilon$, satisfies
\begin{equation}
\epMu(s_1) = \epMu(s_2) \rightarrow \forall_{a} \left|\frac{Q_G(s_1,a)}{\sum_b Q_G(s_1,b)} - \frac{Q_G(s_1,a)}{\sum_b Q_G(s_1,b)}\right| \leq \varepsilon.
\end{equation}
\edefn

\noindent We also assume that the difference in normalizing terms is bounded by some non-negative constant, $k_{\text{mult}} \in \mathbb{R}$, of $\varepsilon$:
\begin{equation}
\left |\sum_i Q_G(s_1,a_i) - \sum_j Q_G(s_2,a_j) \right | \leq k_{\text{mult}} \times \varepsilon.
\end{equation}
\begin{lma} When $S_A$ is created using a function of the $\epMu$ type, for some non-negative constant $k_{\text{mult}} \in \mathbb{R}$:
\begin{equation}
\forall_{s \in S_M} V_G^{\pi^*_G}(s) - V_G^{\pi_{GA}}(s) \leq \frac{2\eps \left(\frac{|\mc{A}|}{1-\gamma} + k_{\text{mult}}\right)}{(1-\gamma)^2}
\end{equation}
\label{lma:mult_lemma}
\end{lma}
\vspace{-7mm}
\noindent {\bf Proof of Lemma~\ref{lma:mult_lemma}} \\

\noindent The proof follows an identical strategy to that of Lemma~\ref{lma:bolt_lemma}, but without the approximation $e^x \approx 1+x$. \qed

%% file: example_domains.tex
\section{Example Domains}
\label{sec:example_domains}

We apply approximate abstraction to five example domains---NChain, Upworld, Taxi, Minefield and Random. These domains were selected for their diversity---NChain is relatively simple, Upworld is particularly illustrative of the power of abstraction, Taxi is goal-based and hierarchical in nature, Minefield is stochastic, and Random MDP has many near-optimal policies.

Our code base\footnote{\url{https://github.com/david-abel/state_abstraction}} provides implementations for abstracting arbitrary \acp{MDP} as well as visualizing and evaluating the resulting abstract \acp{MDP}. We use the graph-visualization library GraphStream~\cite{graphstream} and the planning and \ac{RL} library, BURLAP\footnote{\url{http://burlap.cs.brown.edu/}}. For all experiments, we set $\gamma$ to $0.95$.

\subsection{Visualizations}

We provide visuals of both the ground \ac{MDP} and resulting abstract \ac{MDP} for each domain. A grey circle indicates a state and colored arrows indicate transitions. The thickness of the arrow indicates how much reward is associated with that transition. In the ground \acp{MDP}, states are labeled with a number. In the abstract \acp{MDP}, we indicate which ground states were collapsed to each abstract state by labelling the abstract states with their ground states.

\subsubsection{NChain}

NChain is a simple \ac{MDP} investigated in the Bayesian \ac{RL} literature due to the interesting exploration problem it poses~\cite{dearden1998bayesian}. In our implementation, we set $N=10$, normalized rewards between $0$ and $1$, and used a slip probability of $0.2$. An NChain instance ($N=10$) and its abstraction are visualized in Figure\ref{fig:vis-chain-up}.

In all states, the agent has two actions available: advance down the chain, or return to state 0. The agent receives $.2$ reward for returning to state 0, and no reward for advancing down the chain. The exception is that when the agent transitions to the last state in the chain, it receives $1.0$ reward. Transitions also have small slip probability $\rho$, such that the applied action results in the opposite dynamics. In our implementation, we set $N=10$ and $\rho=0.2$.

\subsubsection{Upworld}

\begin{figure}[b]
\centering
\subfigure[Ground NChain]{
\includegraphics[width=0.20\columnwidth]{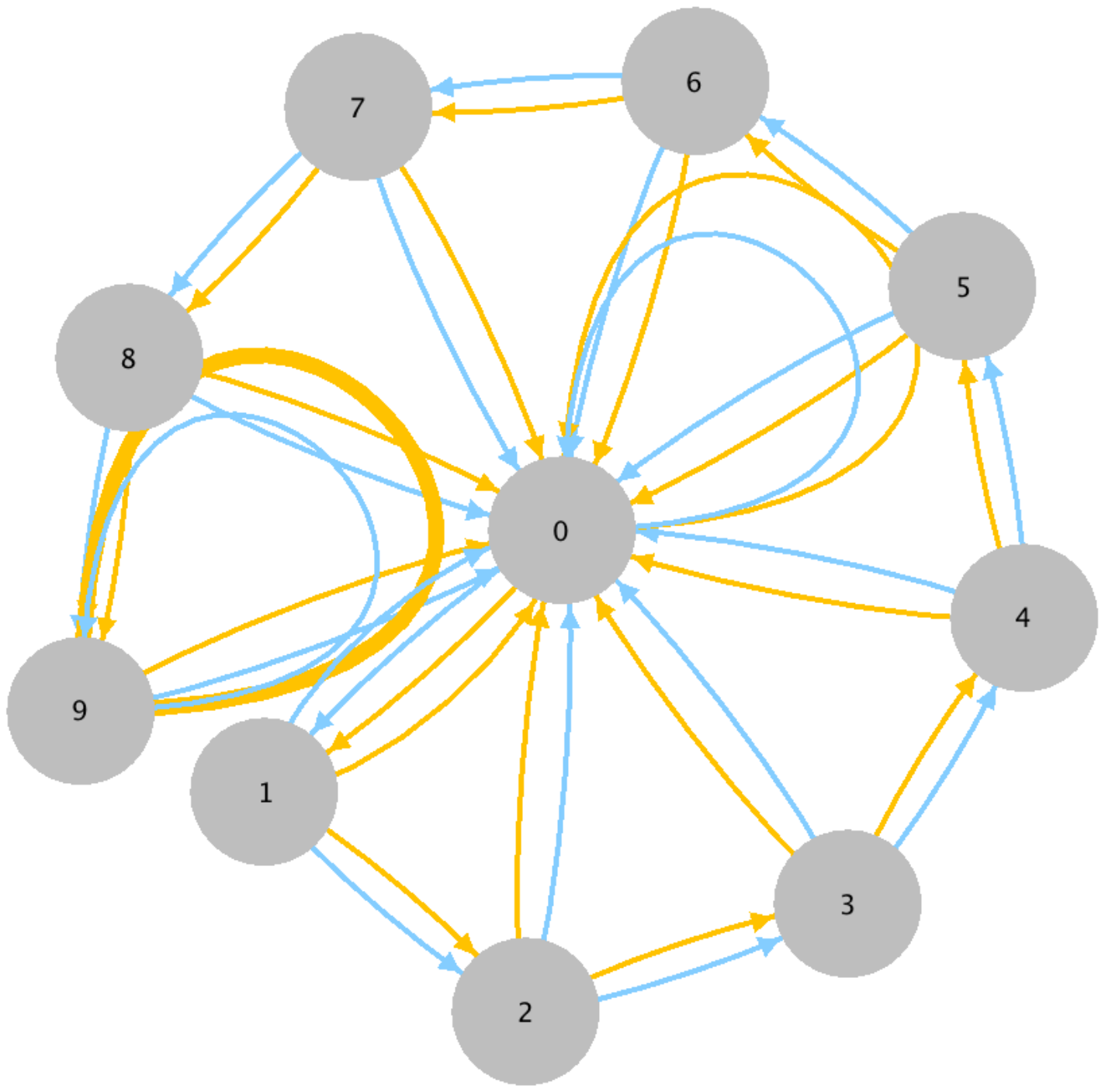}}
\hspace{6mm}
\subfigure[Abstract NChain]{
\includegraphics[width=0.20\columnwidth]{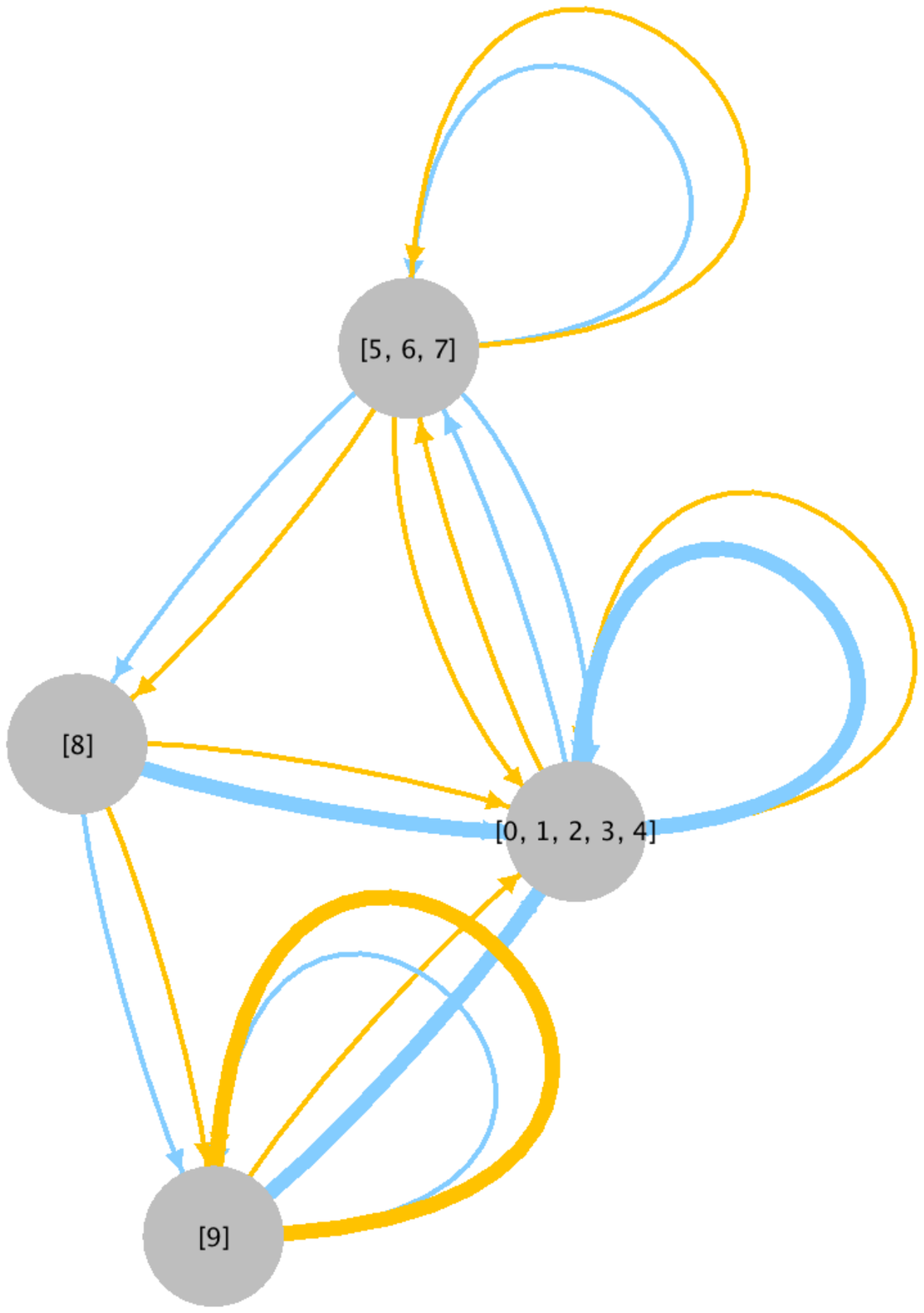}}
\label{fig:nchain-visual}
\subfigure[Ground Upworld]{
\includegraphics[width=0.25\columnwidth]{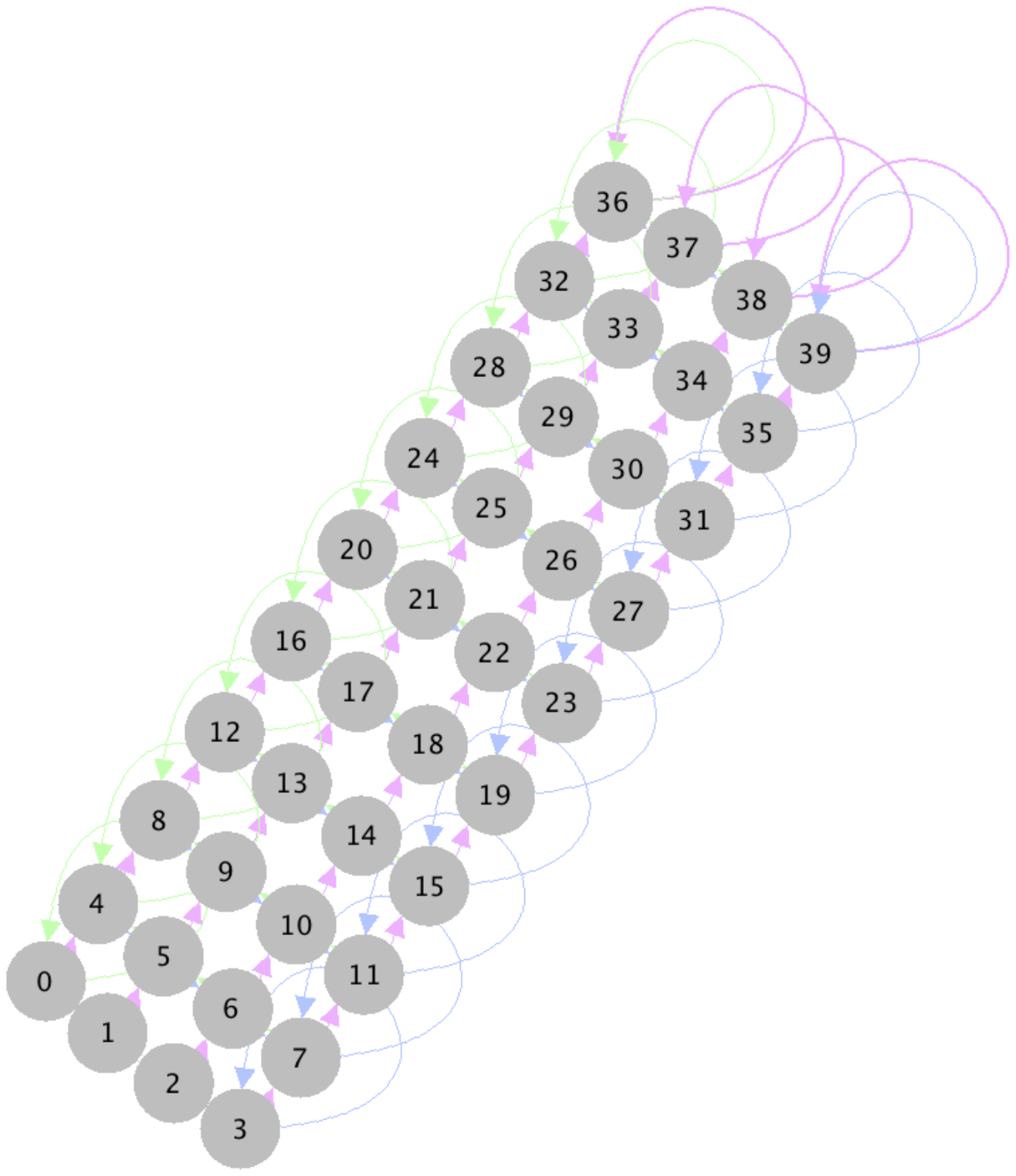}}
\hspace{6mm}
\subfigure[Abstract Upworld]{
\includegraphics[width=0.14\columnwidth]{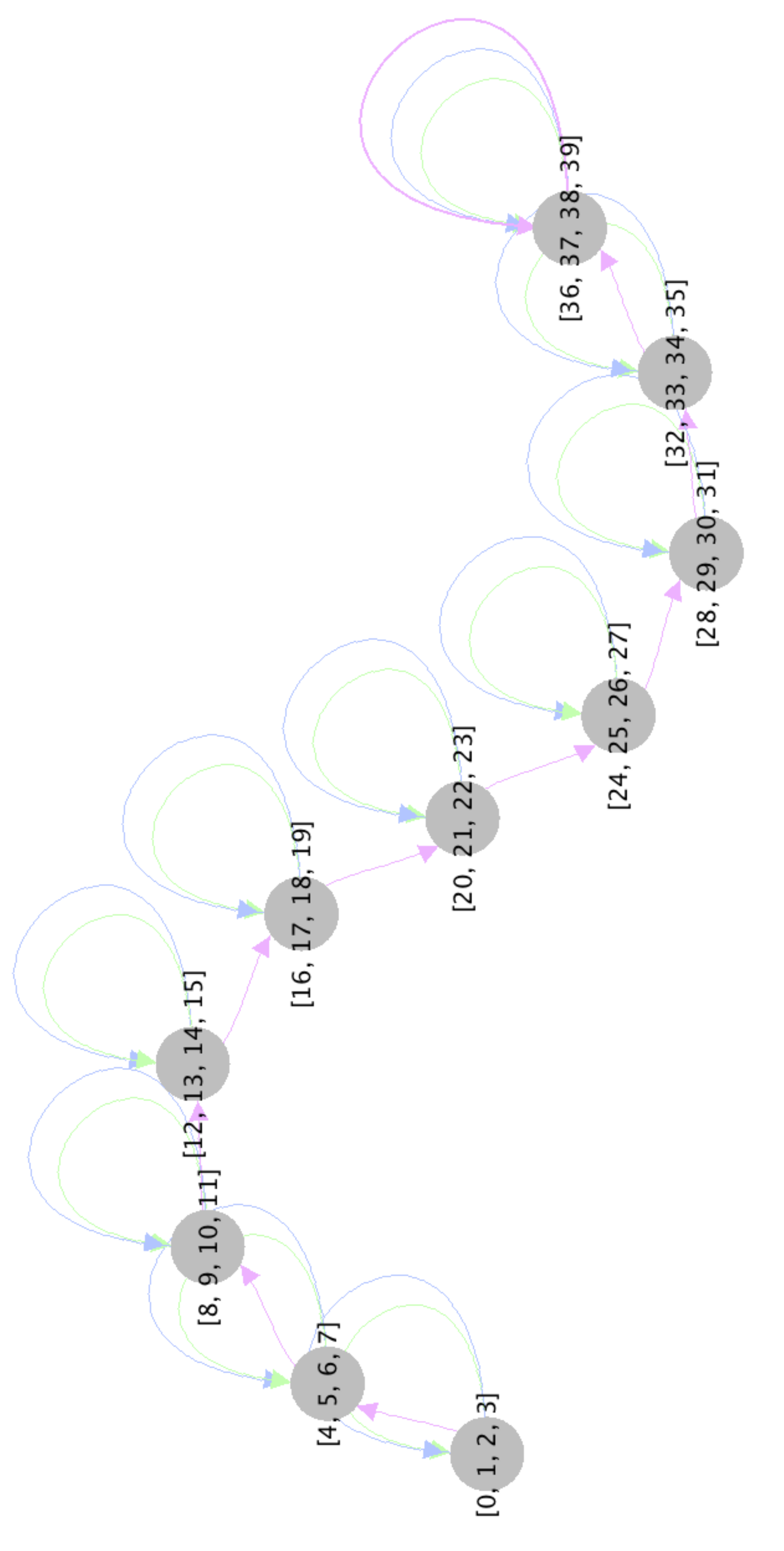}}
\label{fig:vis-chain-up}
\caption{Comparison of the ground and abstract MDPs, under $\epQ$, with $\varepsilon=0.5$}
\end{figure}

The Upworld task is an $N\times M$ grid in which the agent starts in the lower left corner. The agent may move left, right, and up. The agent receives positive reward for transitioning to any state at the top of the grid, where moving up in the top cells self transitions. the agent receives 0 reward for all other transitions. Consequently, moving up is always the optimal action, since moving left and right does not change the agent's manhattan distance to positive reward. During experimentation, we set $N=10$, $M=4$. An Upworld instance ($N=10$, $M=4$) and its abstraction are visualized in Figure\ref{fig:vis-chain-up}.

Upworld illustrates a compelling property with respect to state abstraction: the optimal {\it exact} $Q^*$ abstraction function (when $\eps=0$) can always construct an abstract MDP with $|\mc{S}_A| = N$, the height of the grid, with no change in the value of the optimal policy. Consequently, letting $M$ be arbitrarily large, Upworld offers an arbitrary reduction in the size of the MDP through abstraction, at no cost to the value of the optimal policy. This is a result of the property that all states in the same row have the same $Q$ values: \\

\noindent {\bf Remark}: {\it The optimal exact abstraction, $\phi_{Q^*,0}$, induces an abstract MDP with an optimal policy of equal value to the true optimal policy, and reduces the size of the state space from $N \times M$ (ground) to $N$ (abstract).}

\subsubsection{Taxi}

Taxi has long been studied by the hierarchical RL literature~\cite{dietterich2000hierarchical}. The agent, operating in a Grid World style domain~\cite{russell1995modern}, may move left, right, up, and down, as well as pick up a passenger and drop off a passenger. The goal is achieved when the agent has taken all passengers to their destinations.

\begin{figure}[t]
\centering
\subfigure[Ground Taxi]{
\includegraphics[width=0.25\columnwidth]{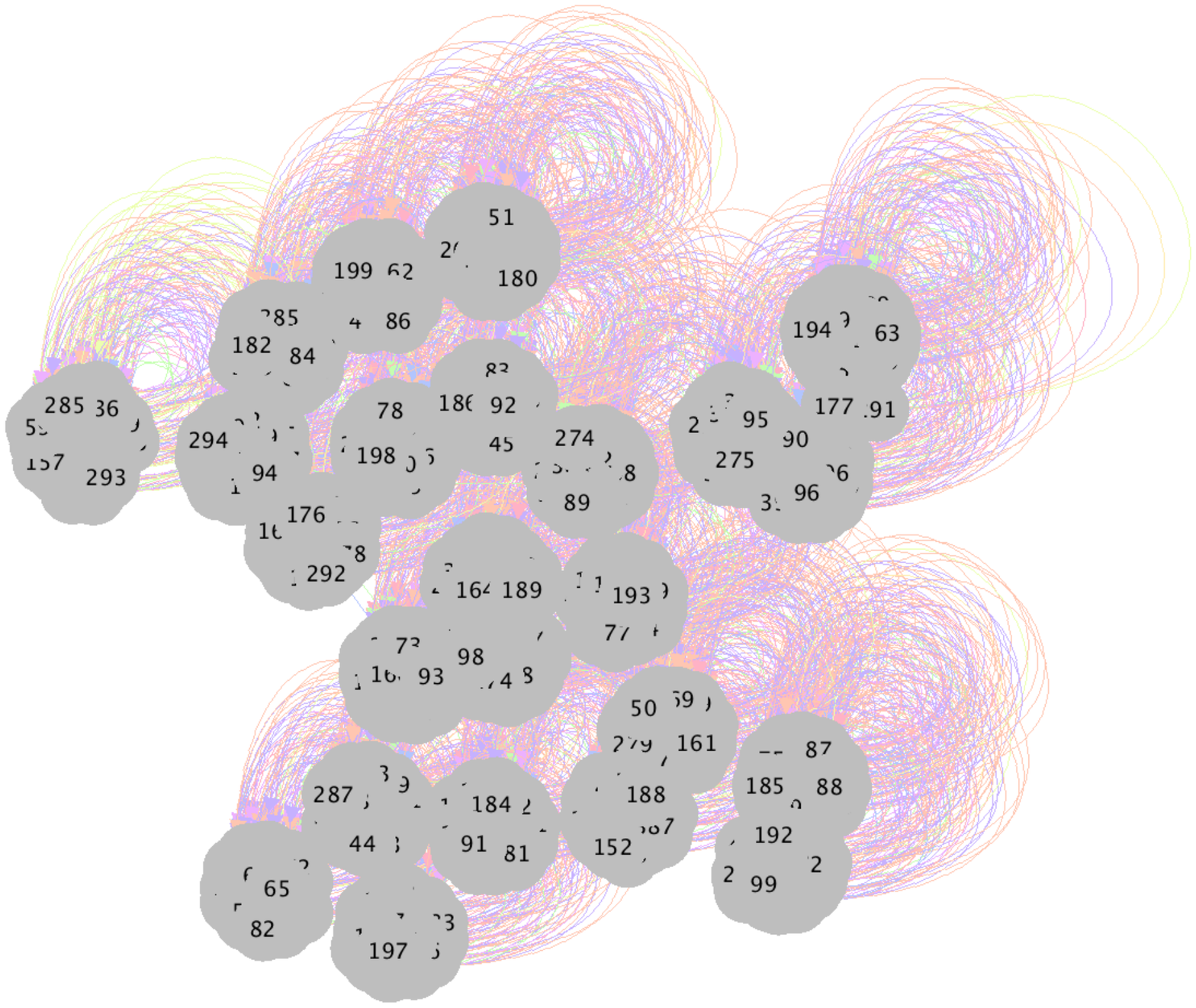}}
\hspace{6mm}
\subfigure[Abstract Taxi]{
\includegraphics[width=0.25\columnwidth]{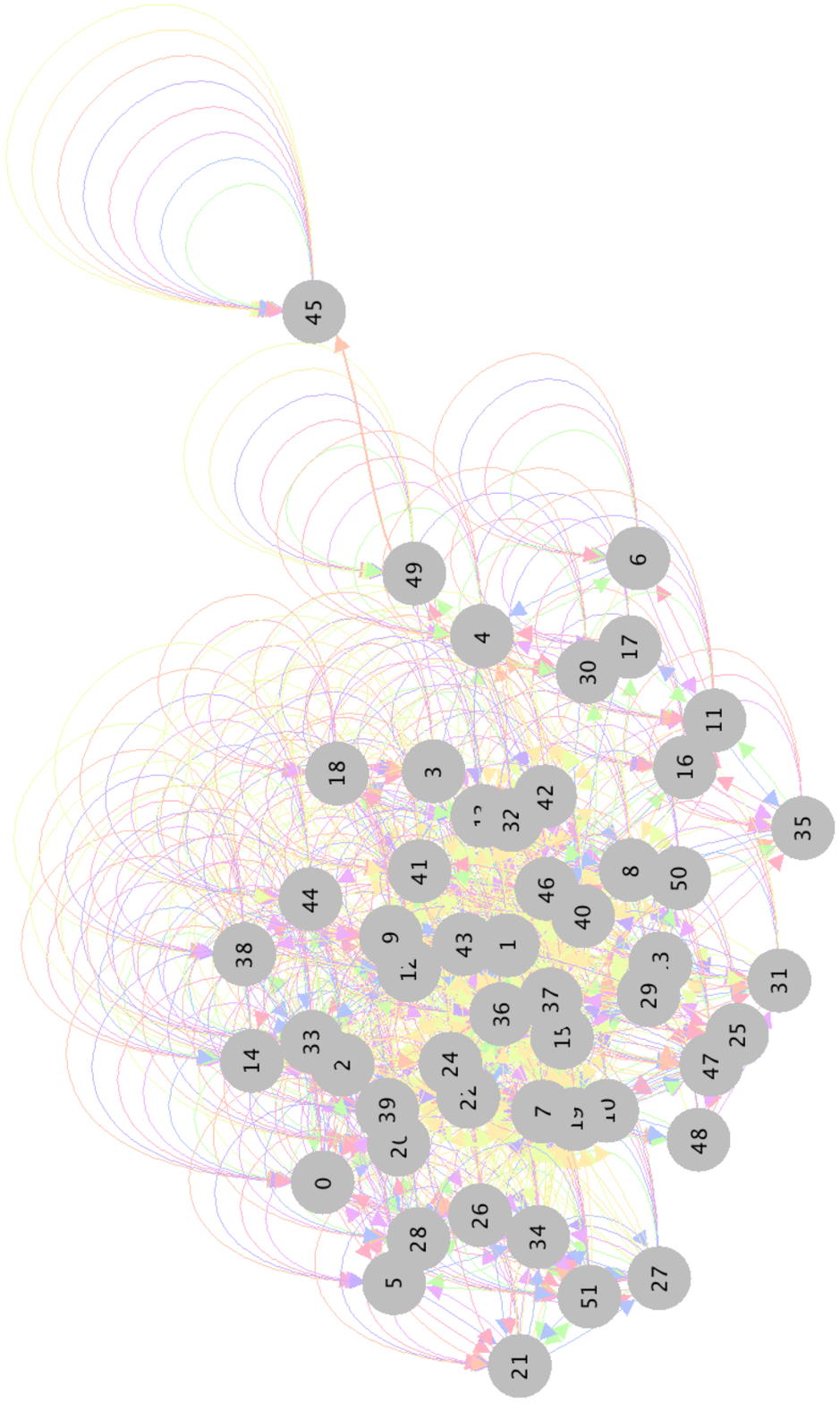}}
\caption{Comparison of the ground and abstract Taxi MDPs under an $\epQ$ abstraction, with $\varepsilon=0.03$.}
\label{fig:vis-taxi}
\end{figure}

We visualize the compression on a simple 626 Taxi instance in Figure~\ref{fig:vis-taxi}. As stated above, we visualize the original Taxi problem into a graph representation so that we may visualize both the ground MDP and abstract MDP in the same format, despite the unnatural appearance.

\subsubsection{Minefield}
Minefield is a test problem we are introducing that uses the Grid World dynamics of \citet{russell1995modern} with slip probability of $x$. The reward function is such that moving up in the top row of the grid receives $1.0$ reward; all other transitions receive $0.2$ reward, except for transitions to a random set of $\kappa$ mine-states (which may include the top row) that receive $0$ reward. We set $N=10, M=4, \varepsilon=0.5, \kappa = 5, x = 0.01$.

\begin{figure}[b]
\centering
\subfigure[Ground Minefield]{
\includegraphics[width=0.15\columnwidth]{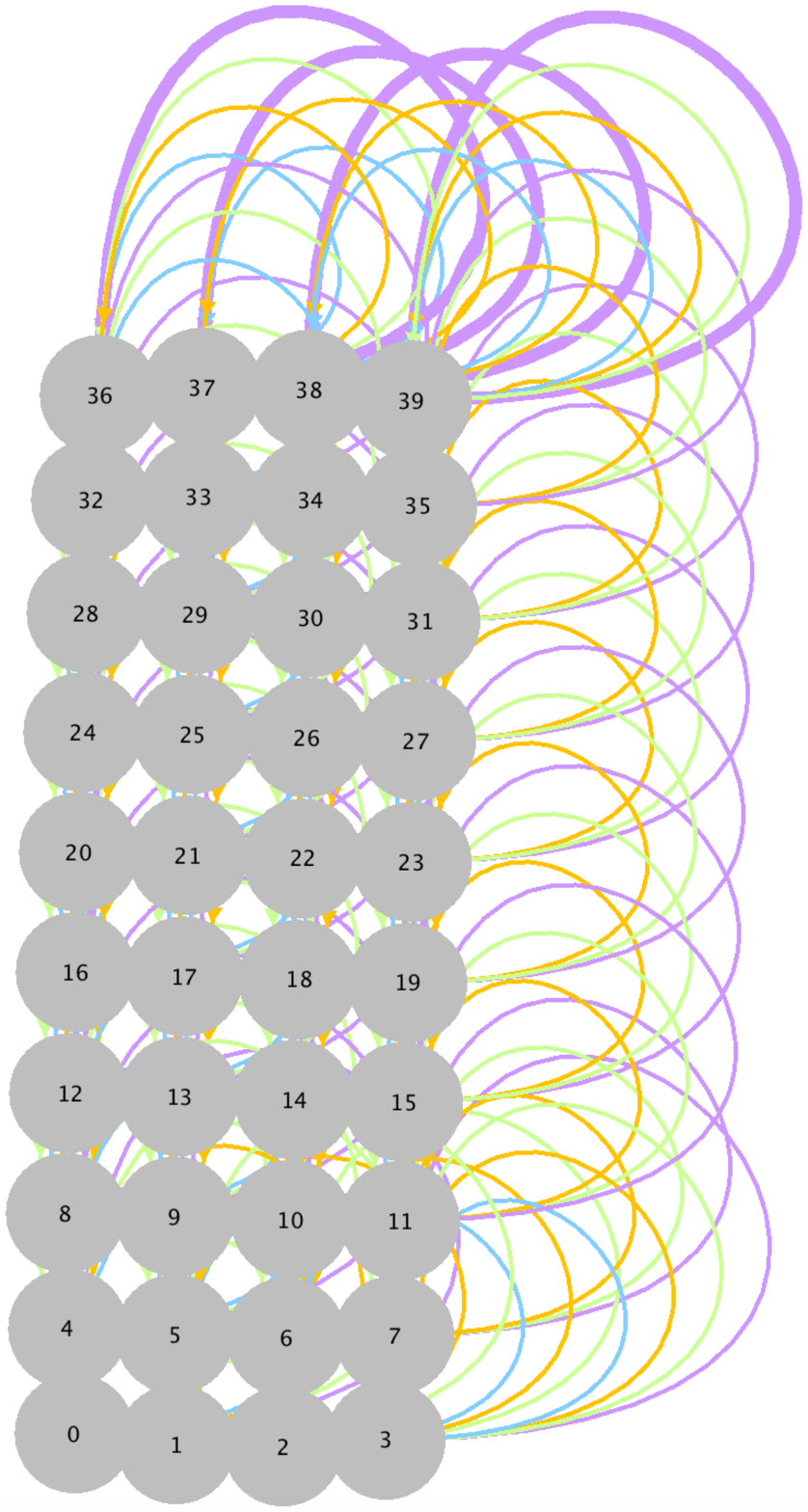}}
\hspace{6mm}
\subfigure[Abstract Minefield]{
\includegraphics[width=0.13\columnwidth]{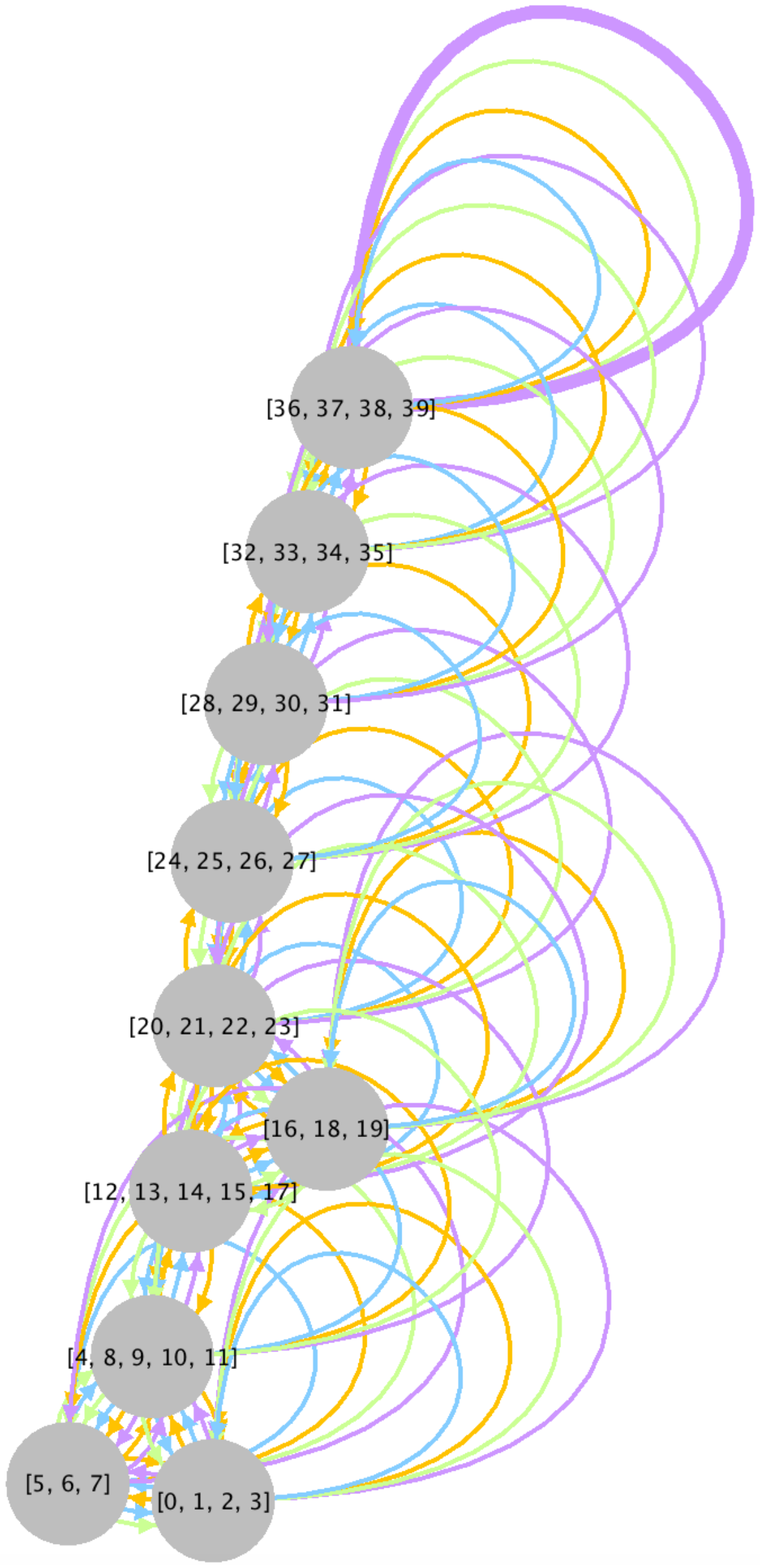}}
\subfigure[Ground Random]{
\includegraphics[width=0.25\columnwidth]{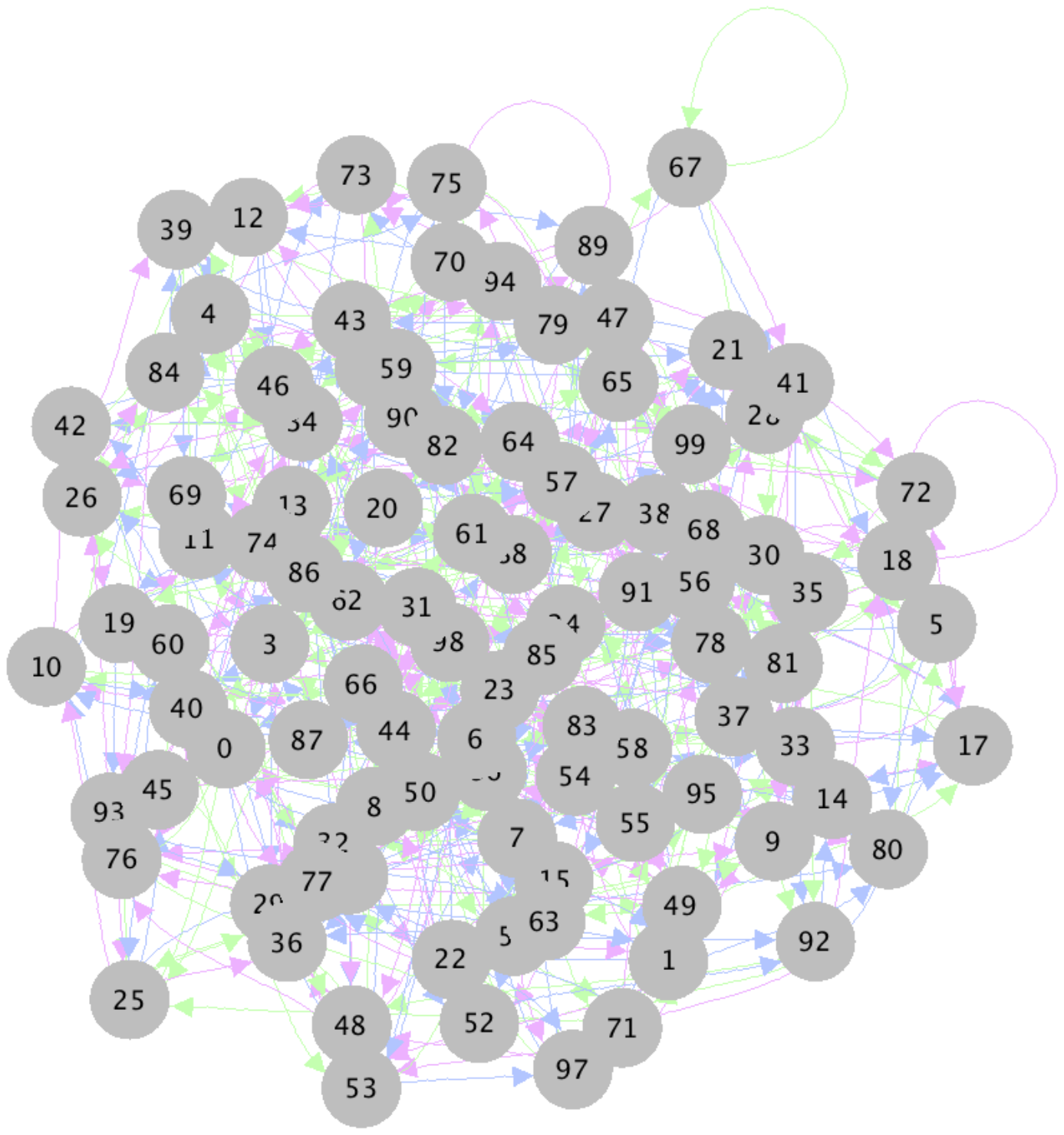}}
\hspace{6mm}
\subfigure[Abstract Random]{
\includegraphics[width=0.25\columnwidth]{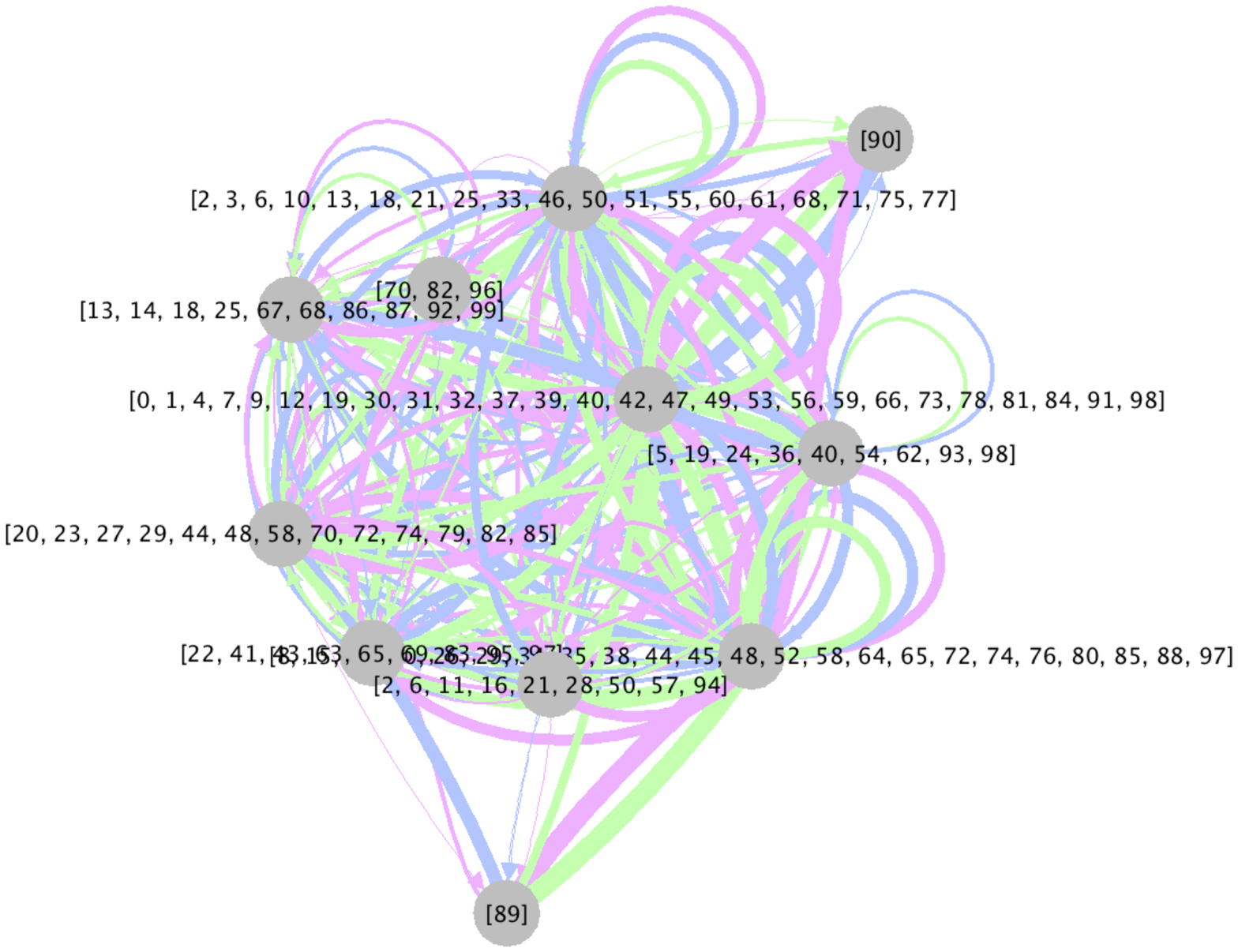}}
\caption{Comparison of the ground and abstract MDPs under an $\epQ$ abstraction, with $\varepsilon=0.5$.}
\label{fig:vis-mine-rand}
\end{figure} 

\subsubsection{Random MDP}
In the Random \ac{MDP} domain we consider, there are $100$ states and $3$ actions. For each state, each action transitions to one of two randomly selected states with probability $0.5$. The Random MDP and its compression are visualized in Figure~\ref{fig:vis-mine-rand}.


%% file: results_discussion.tex
\section{Empirical Results}
\label{sec:results}

We ran experiments on the $\epQ$ type aggregation functions. We provide results for only $\epQ$ because, as our proofs in Section \ref{sec:theory_results} demonstrate, the other three functions are reducible to particular $\epQ$ functions. For the purpose of illustrating what kinds of approximations are possible we built each abstraction by first solving the MDP, then greedily aggregating ground states into abstract states that satisfied the $\epQ$ criteria. Since this approach represents an order-dependent approximation to the maximum amount of abstraction possible, we randomized the order in which states were considered across trials. Every ground state is equally weighted in its abstract state.

\begin{figure*}[h]
\centering
\subfigure[NChain]{
\includegraphics[width=0.26\columnwidth]{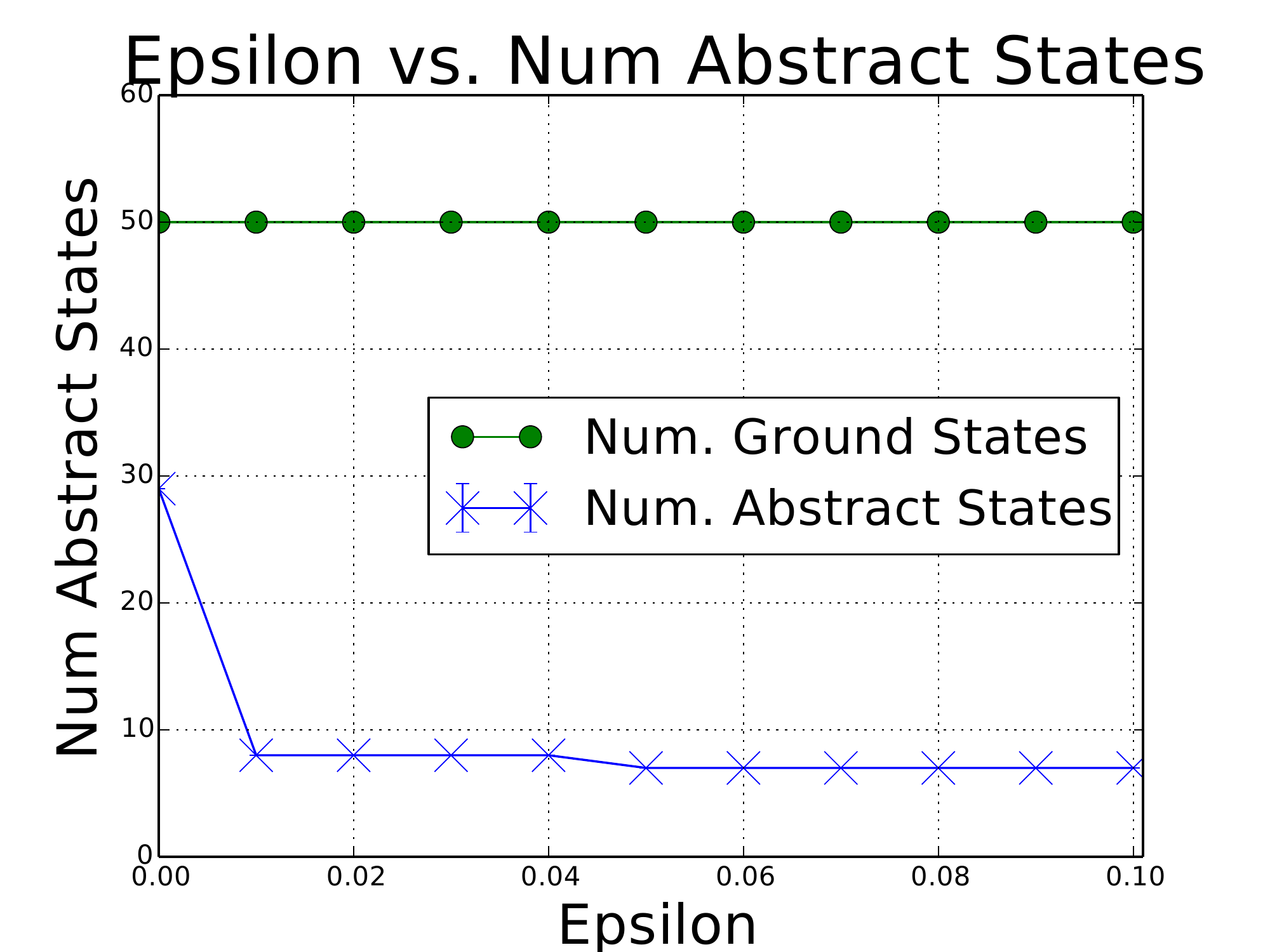}
}\hspace{8mm}
\subfigure[NChain]{
\includegraphics[width=0.26\columnwidth]{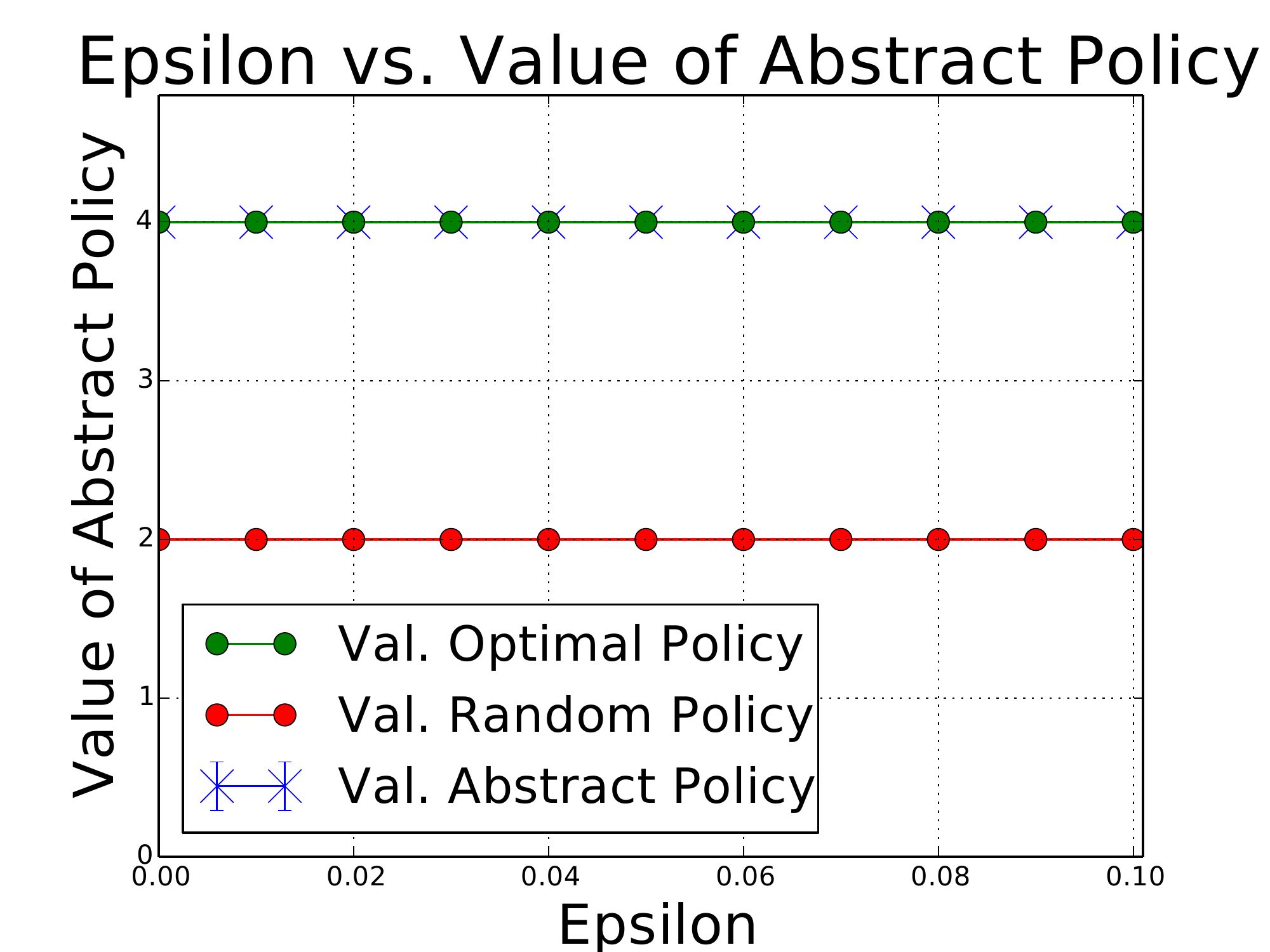}
} \\
\subfigure[Upworld]{
\includegraphics[width=0.26\columnwidth]{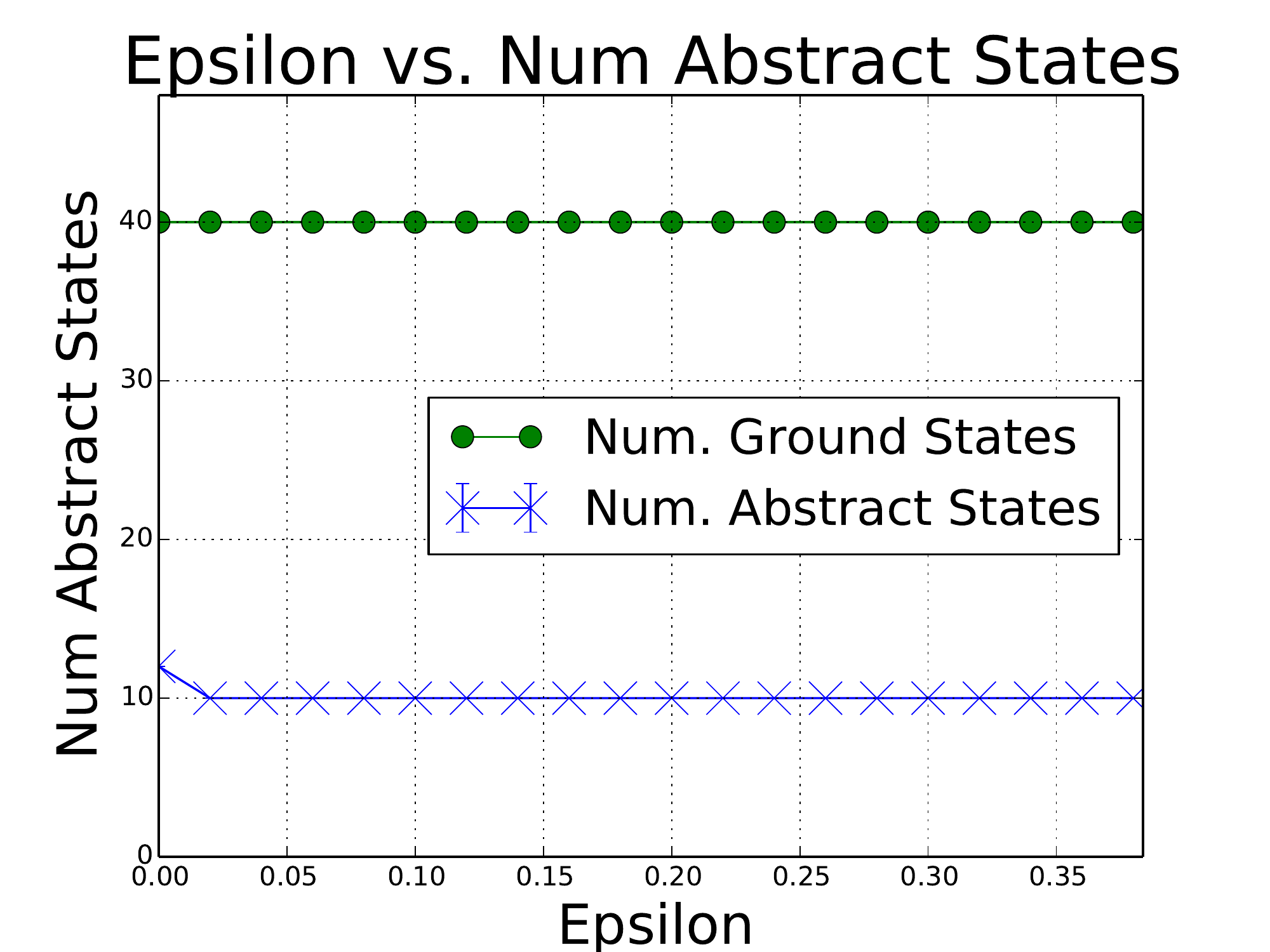}
}\hspace{8mm}
\subfigure[Upworld]{
\includegraphics[width=0.26\columnwidth]{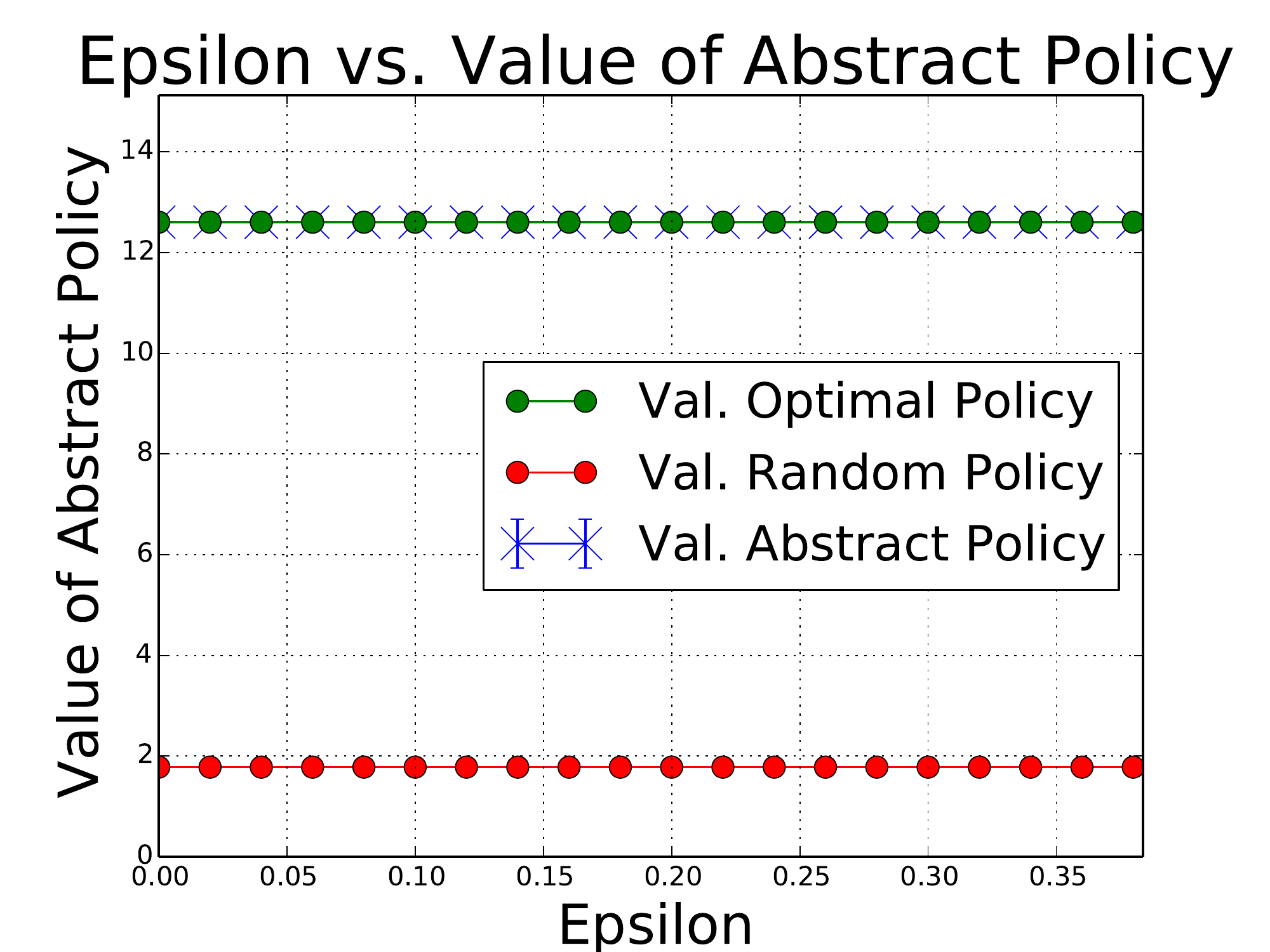}
}
\caption{$\varepsilon$ vs. Num States (left) and $\varepsilon$ vs. Abstract Policy Value (right).\label{fig:first_empirical_results}}
\end{figure*}

\begin{figure*}
\centering
\subfigure[Minefield]{
\includegraphics[width=0.26\columnwidth]{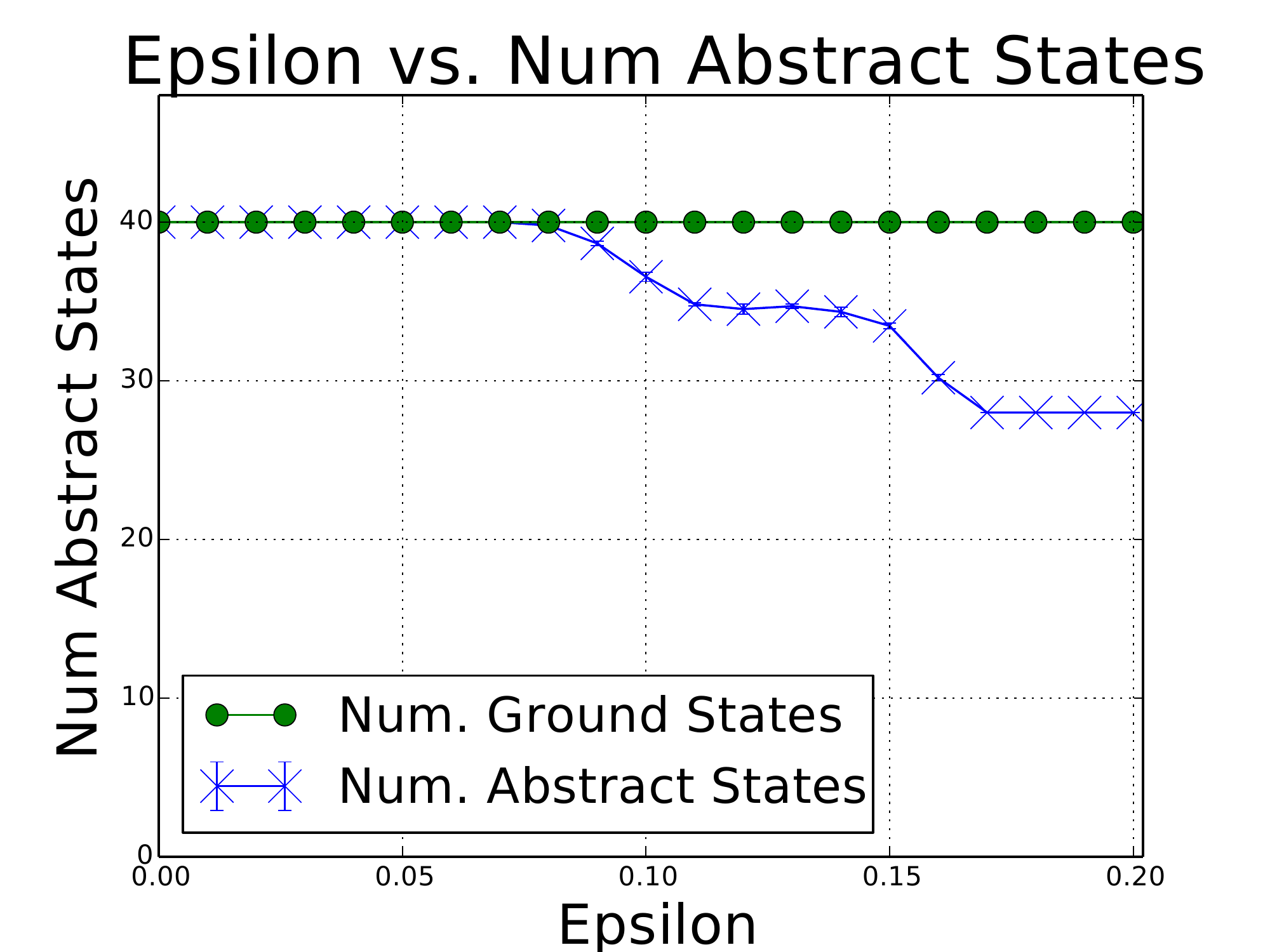}
}\hspace{8mm}
\subfigure[Minefield]{
\includegraphics[width=0.26\columnwidth]{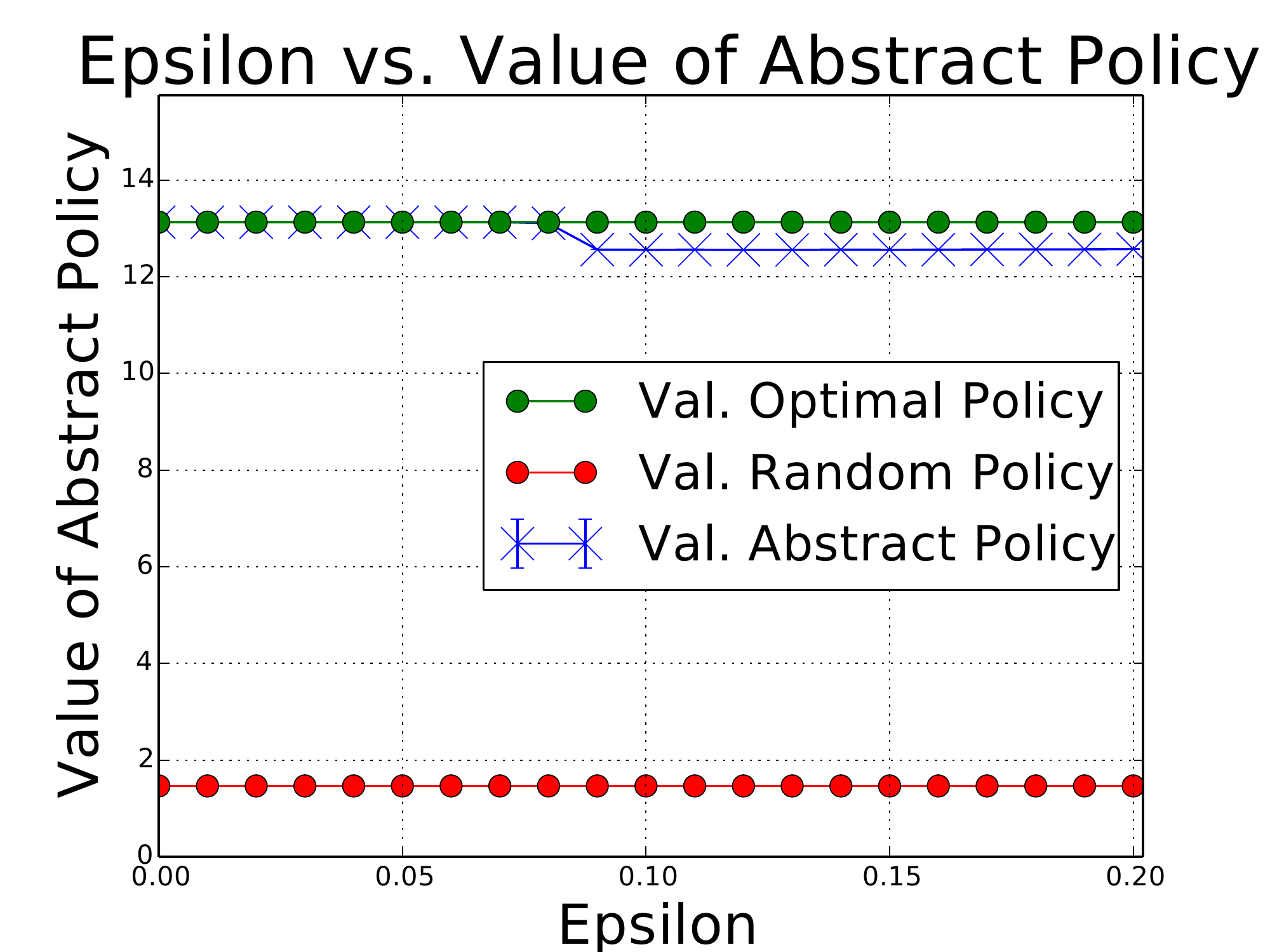}
} \\
\subfigure[Taxi]{
\includegraphics[width=0.26\columnwidth]{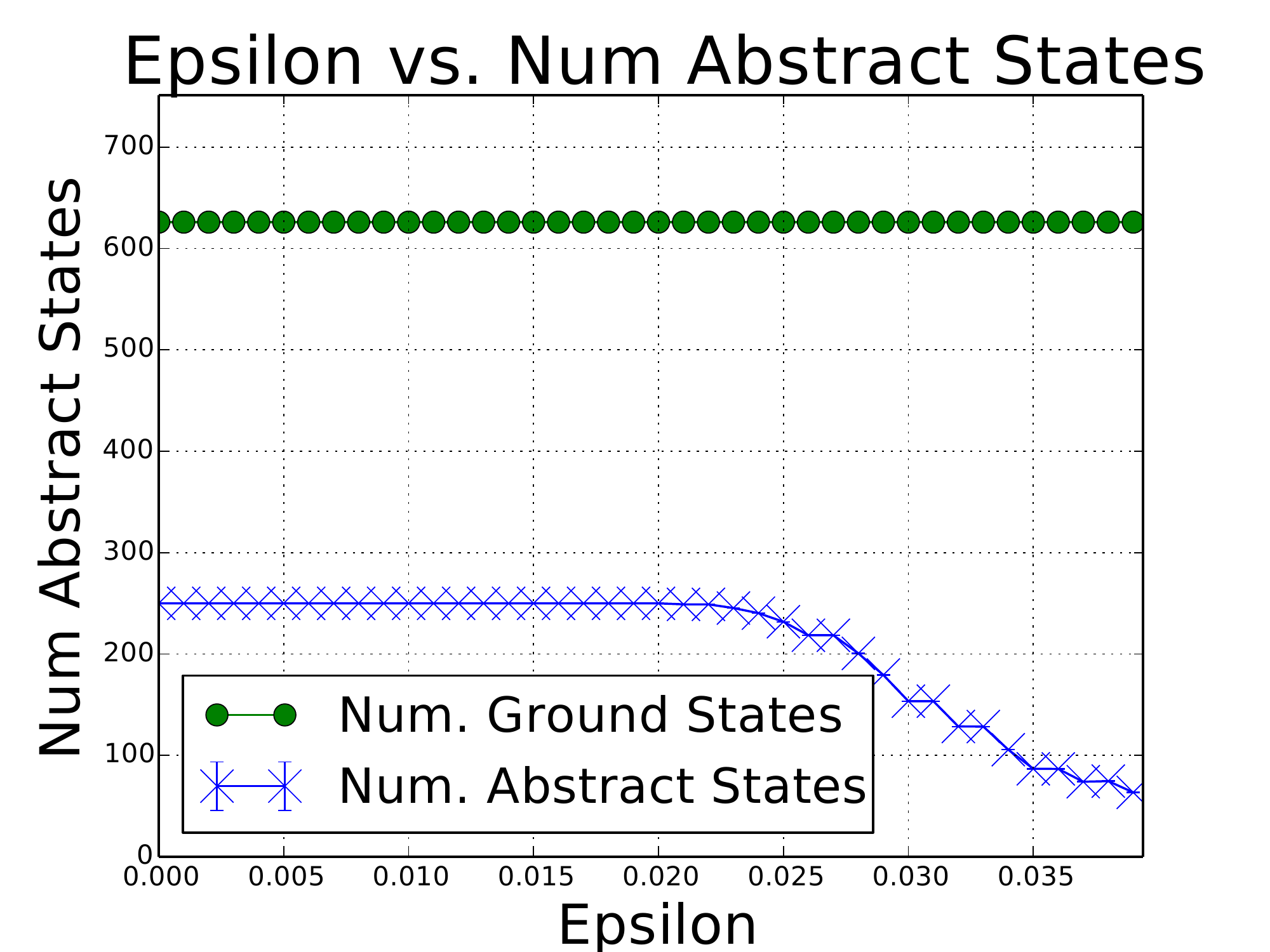}
}\hspace{8mm}
\subfigure[Taxi]{
\includegraphics[width=0.26\columnwidth]{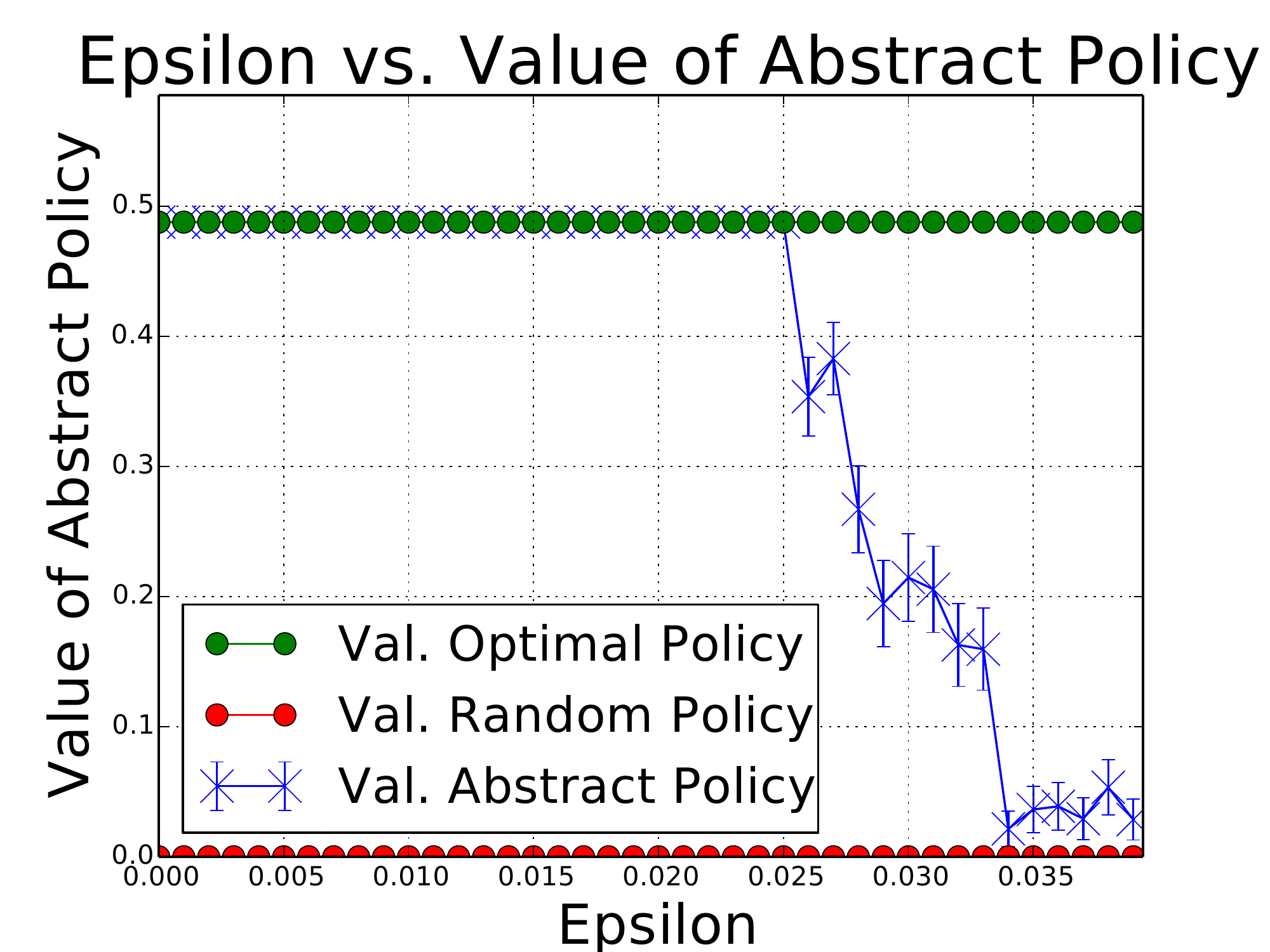}
} \\
\subfigure[Random]{
\includegraphics[width=0.26\columnwidth]{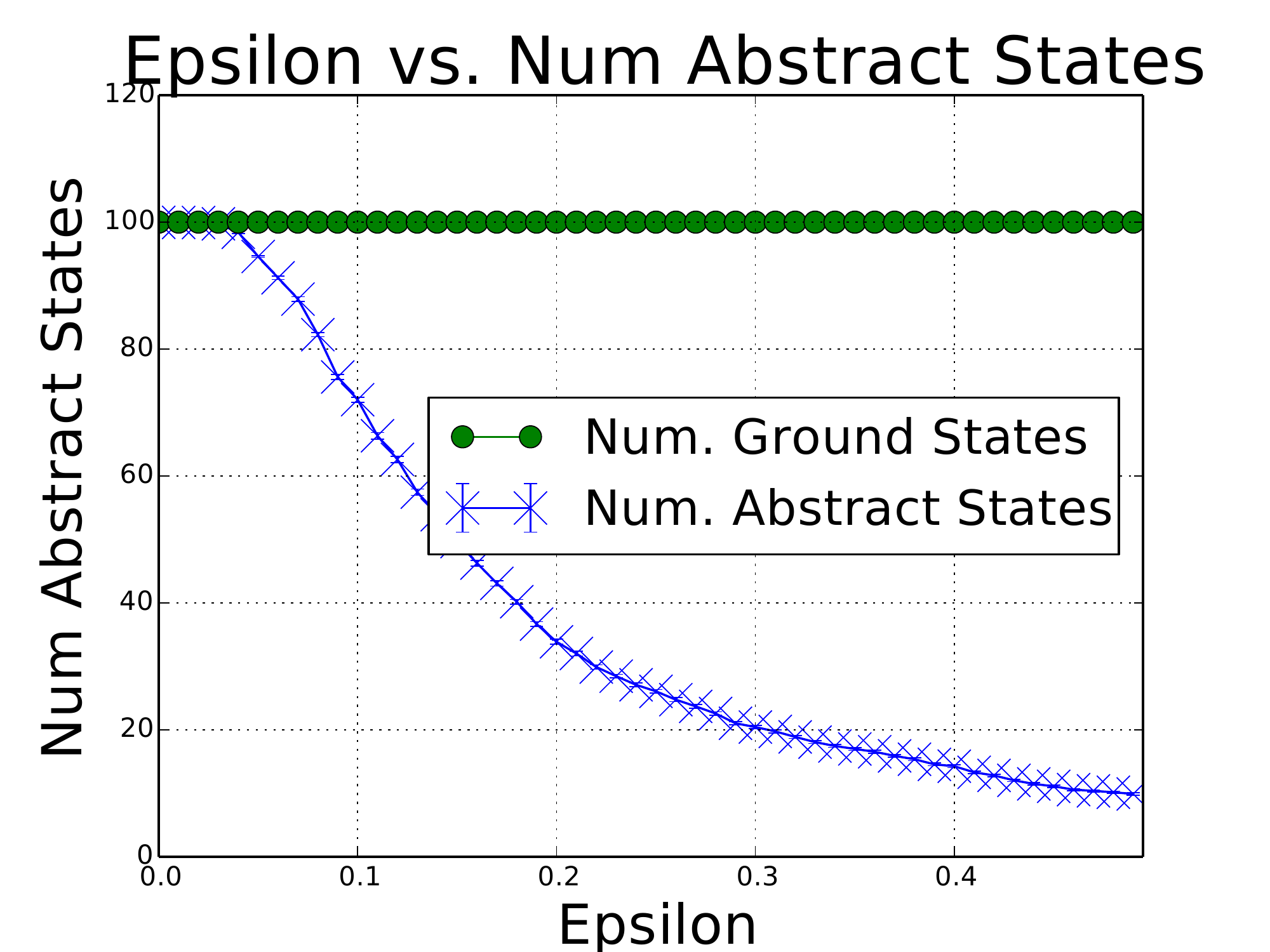}
}\hspace{8mm}
\subfigure[Random]{
\includegraphics[width=0.26\columnwidth]{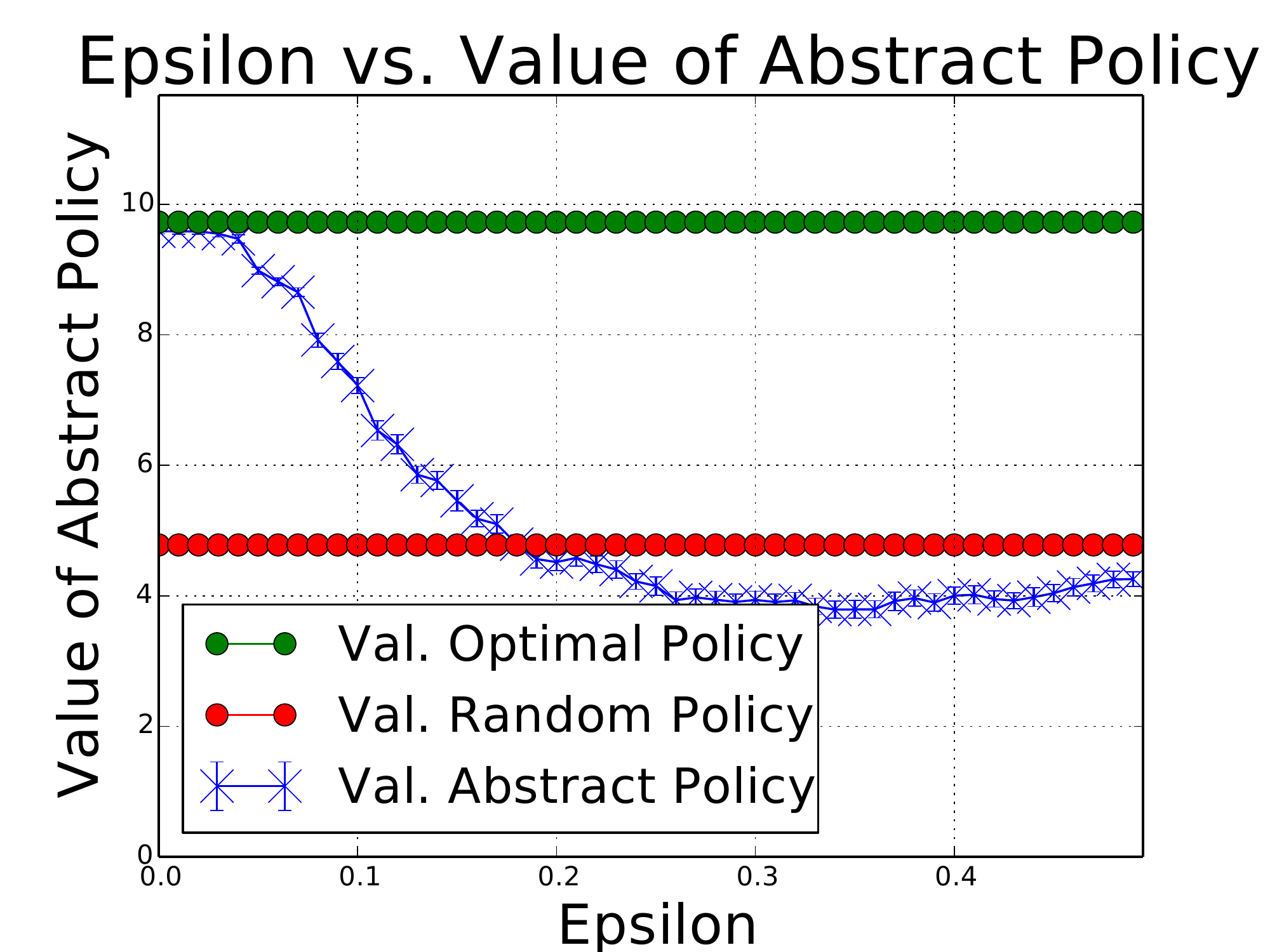}
}
\caption{$\varepsilon$ vs. Num States (left) and $\varepsilon$ vs. Abstract Policy Value (right).\label{fig:sec_empirical_results}}
\end{figure*}

For each domain, we report two quantities as a function of epsilon with 95\% confidence bars. First, we compare the number of states in the abstract \ac{MDP} for different values of $\varepsilon$, shown in the left column of Figure~\ref{fig:first_empirical_results} and Figure~\ref{fig:sec_empirical_results}. The smaller the number of abstract states, the smaller the state space of the $\ac{MDP}$ that the agent must plan over. Second, we report the value under the abstract policy of the initial ground state, also shown in the right column of Figure~\ref{fig:first_empirical_results} and Figure~\ref{fig:sec_empirical_results}. In the Taxi and Random domains, 200 trials were run for each data point, whereas 20 trials were sufficient in Upworld, Minefield, and NChain.

Our empirical results corroborate our thesis---approximate state abstractions can decrease state space size while retaining bounded error. In both NChain and Minefield, we observe that, as $\varepsilon$ increases from $0$, the number of states that must be planned over is reduced, and optimal behavior is either fully maintained (NChain) or very nearly maintained (Minefield). Similarly for Taxi, when $\varepsilon$ is between $.02$ and $.025$, we observe a reduction in the number of states in the abstract \ac{MDP} while value is fully maintained. After $.025$, increased reduction in state space size comes at a cost of value. Lastly, as $\varepsilon$ is increased in the Random domain, there is a smooth reduction in the number of abstract states with a corresponding cost in the value of the derived policy. When $\varepsilon = 0$, there is no reduction in state space size whatsoever (the ground \ac{MDP} has 100 states), because no two states have identical optimal $Q$-values.

Our experimental results also highlight a noteworthy characteristic of approximate state abstraction in goal-based \acp{MDP}. Taxi exhibits relative stability in state space size and behavior for $\varepsilon$ up to $.02$, at which point both fall off dramatically. We attribute the sudden fall off of these quantities to the goal-based nature of the domain; once information critical for achieving optimal behavior is lost in the state aggregation, solving the goal---and so acquiring any reward---is impossible. Conversely, in the Random domain, a great deal of near optimal policies are available to the agent. Thus, even as the information for optimal behavior is lost, there are many near optimal policies available to the agent that remain available.

%% file: conclusion.tex
\section{Conclusion}

Approximate abstraction in \acp{MDP} offers considerable advantages over exact abstraction. First, approximate abstraction relies on criteria that we imagine a planning or learning algorithm to be able to learn without solving the full \ac{MDP}. Second, approximate abstractions can achieve greater degrees of compression due to their relaxed criteria of equality. Third, methods that employ approximate aggregation techniques are able to tune the aggressiveness of abstraction all the while incurring bounded error. In this work, we proved bounds for the value lost when behaving according to the optimal policy of the abstract \ac{MDP}, and empirically demonstrate that approximate abstractions can reduce state space size with minor loss in the quality of the behavior. We provide a code base that provides implementations to abstract, visualize, and evaluate an arbitrary MDP to promote further investigation into approximate abstraction.


There are many directions for future work.
First, we are interested in extending the approach of \citet{ortner2013adaptive} by learning the approximate abstraction functions introduced in this paper online in the planning or \ac{RL} setting, particularly when the agent must solve a collection of related MDPs.
Additionally, while our work presents several sufficient conditions for achieving bounded error of learned behavior with approximate abstractions, we hope to investigate what conditions are strictly necessary for an approximate abstraction to achieve bounded error.
Further, we are interested in characterizing the relationship between temporal abstractions, such as options~\cite{sutton1999between} and approximate state abstractions.
Lastly, we are interested in understanding the relationship between various approximate abstractions and the information theoretical limitations on the degree of abstraction achievable in \acp{MDP}.
